\documentclass{article}

\PassOptionsToPackage{numbers, compress}{natbib}
\usepackage[main, final]{neurips_2026}

\usepackage[utf8]{inputenc} 
\usepackage[T1]{fontenc}    

\usepackage{url}            
\usepackage{booktabs}       
\usepackage{amsfonts}       
\usepackage{nicefrac}       
\usepackage{microtype}      
\usepackage{xcolor}         
\definecolor{icmlblue}{HTML}{DCEBFF}
\definecolor{lightblue}{rgb}{0.21,0.49,0.74}
\usepackage[colorlinks=true, allcolors=lightblue]{hyperref}

\usepackage{amsmath,amsfonts,bm}

\def\eqref#1{equation~\ref{#1}}

\def\1{\bm{1}}

\DeclareMathAlphabet{\mathsfit}{\encodingdefault}{\sfdefault}{m}{sl}
\SetMathAlphabet{\mathsfit}{bold}{\encodingdefault}{\sfdefault}{bx}{n}

\usepackage{graphicx}
\graphicspath{{./}}
\makeatletter
\def\input@path{{./}}
\makeatother
\usepackage{subcaption}
\usepackage{adjustbox}
\usepackage{amsmath}
\usepackage{amssymb}
\usepackage{mathtools}
\usepackage{amsthm}
\usepackage{xspace}
\usepackage{dsfont}

\usepackage{enumitem}
\usepackage{caption}
\usepackage{multirow}
\usepackage{colortbl}
\usepackage{bbm}

\usepackage{setspace}
\usepackage{bm}
\usepackage{soul}
\usepackage{csquotes}

\usepackage{duckuments}
\usepackage{wrapfig}

\usepackage{svg}
\usepackage{pifont}
\usepackage{arydshln}
\usepackage{siunitx}

\newcommand{\cmark}{\ding{51}}%
\newcommand{\xmark}{\ding{55}}%

\usepackage[capitalize,noabbrev]{cleveref}
\usepackage{tcolorbox}

\theoremstyle{plain}

\theoremstyle{definition}

\theoremstyle{remark}

\definecolor{icmlgray}{HTML}{F5F5F9}

\usepackage[textsize=tiny]{todonotes}
\newcommand{\model}{\textsc{HYDRA}\xspace}
\newcommand{\tokenizer}{\textsc{HYDRA-TOK}\xspace}
\newcommand{\g}[1]{\textcolor{gray}{#1}}
\newcommand{\qxr}[1]{{\color{black} #1}}

\newcommand{\huggingface}{\raisebox{-0.25em}{\includegraphics[height=1.2em]{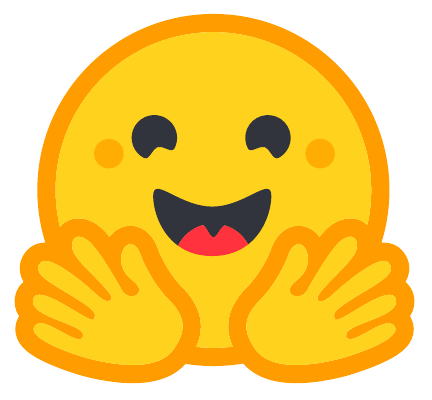}}\xspace}
\newcommand{\github}{\raisebox{-0.25em}{\includegraphics[height=1.2em]{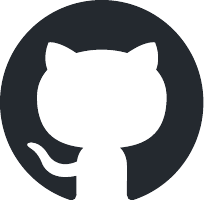}}\xspace}
\newcommand{\wandbicon}{\raisebox{-0.25em}{\includegraphics[height=1.2em]{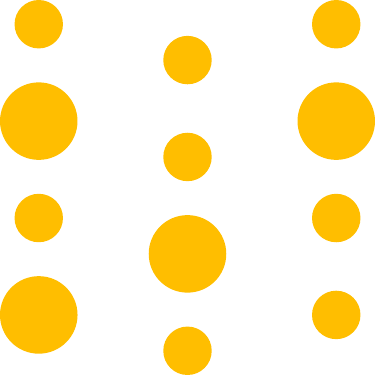}}\xspace}
\newcommand{\webicon}{%
  \raisebox{-0.25em}{\includegraphics[height=1.2em]{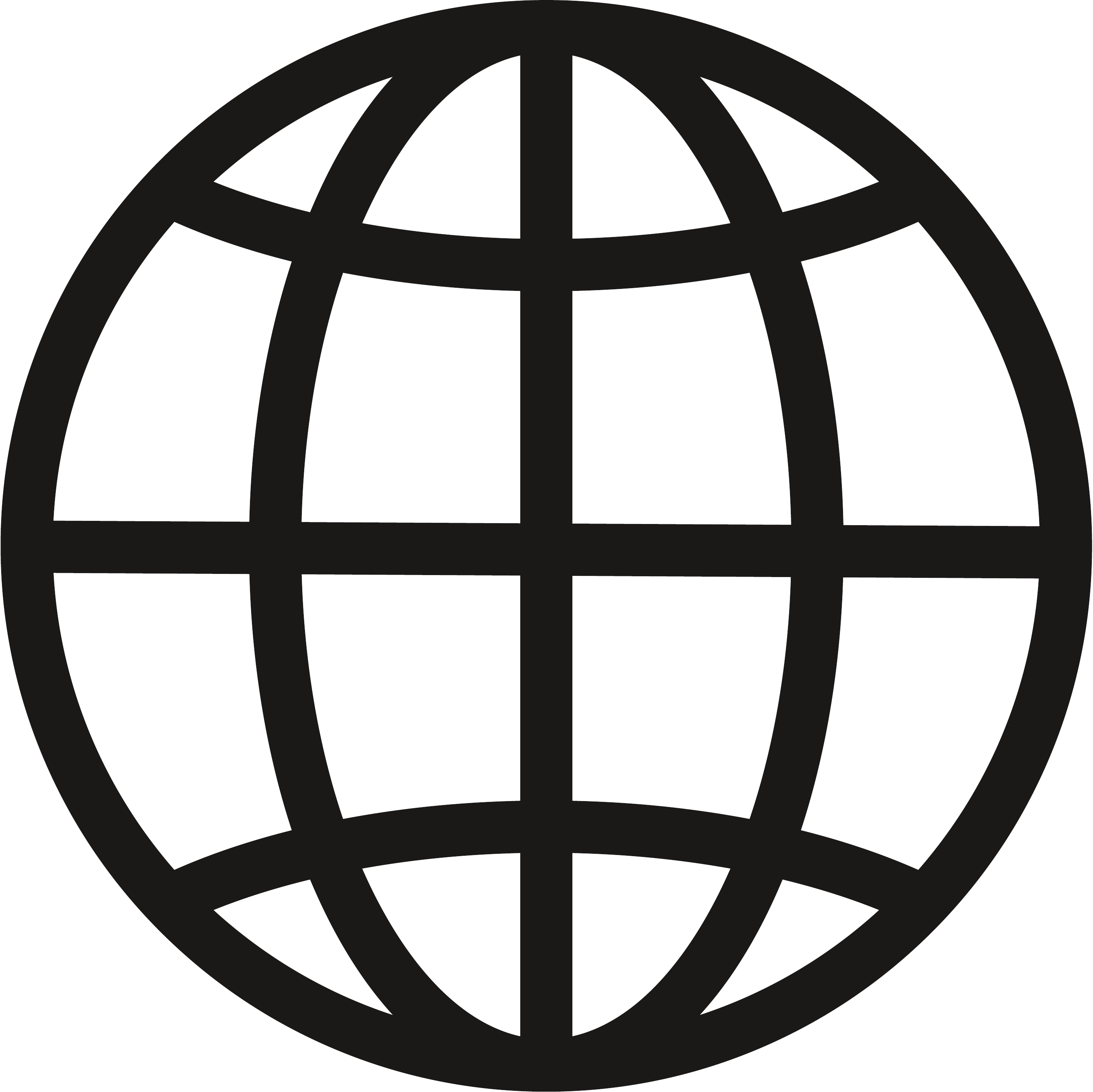}}\xspace
}

\newcommand{\linkbutton}[3]{%
  \href{#1}{%
    \tcbox[
      on line, arc=3pt, colback=white, colframe=white,
      boxrule=0.5pt, size=small,
      left=1mm, right=1mm, top=0.5mm, bottom=0.5mm,
      fontupper=\footnotesize\sffamily
    ]{#2 \textbf{#3}}%
  }%
}

\title{\model:\, Representation Harmonized Tokenization for Multimodal Generation and Understanding}

\author{%
  \normalfont
  \begin{tabular}{@{}c@{}}
    {\small\bfseries
      Xuerui Qiu$^{1,2,3}$\thanks{These authors contributed equally to this work.} \enspace
      Yutao Cui$^{2}$\footnotemark[1] \enspace
      Guozhen Zhang$^{2,4}$\footnotemark[1] \enspace
      Junzhe Li$^{2,5}$ \enspace
      Yaqi Zhao$^{2,5}$ \enspace
      Xiao Zhang$^{2}$}
    \\[0.35em]
    {\small\bfseries
      Yang Li$^{2}$ \enspace
      Songtao Liu$^{2}$ \enspace
      Miles Yang$^{2}$ \enspace
      Yu Shi$^{3}$\thanks{Corresponding authors:
        Yu Shi (\texttt{shiyu@bza.edu.cn}) and
        Zhao Zhong (\texttt{caesarqin@tencent.com}).} \enspace
      Zhao Zhong$^{2}$\footnotemark[2] \enspace
      Liefeng Bo$^{2}$}
    \\[0.7em]
    {\small
      $^{1}$Institute of Automation, Chinese Academy of Sciences
      \quad $^{2}$Tencent Hunyuan}
    \\[0.2em]
    {\small
      $^{3}$Zhongguancun Academy
      \qquad $^{4}$Nanjing University
      \qquad $^{5}$Peking University}
    \\[0.35em]
    \vspace{-5pt}
{\small
  \linkbutton{https://bollossom.github.io/HYDRA/}{\webicon}{Website}
  \hspace{0.8em}
  \linkbutton{https://github.com/Tencent-Hunyuan/HYDRA}{\github}{Code}
  \hspace{0.8em}
  \linkbutton{https://huggingface.co/Xuerui123/HYdra}{\huggingface}{Models}
  \hspace{0.8em}
  \linkbutton{https://bollossom.github.io/HYDRA/training-logs/}{\wandbicon}{Training Logs}
}
  \end{tabular}%
}
\begin{document}

\maketitle

\vspace{-10pt}
\begin{figure}[ht]
\vspace{-20pt}
\begin{center}
    \subcaptionbox{Comparison with SOTA.}
        {\includegraphics[width=0.37\linewidth]{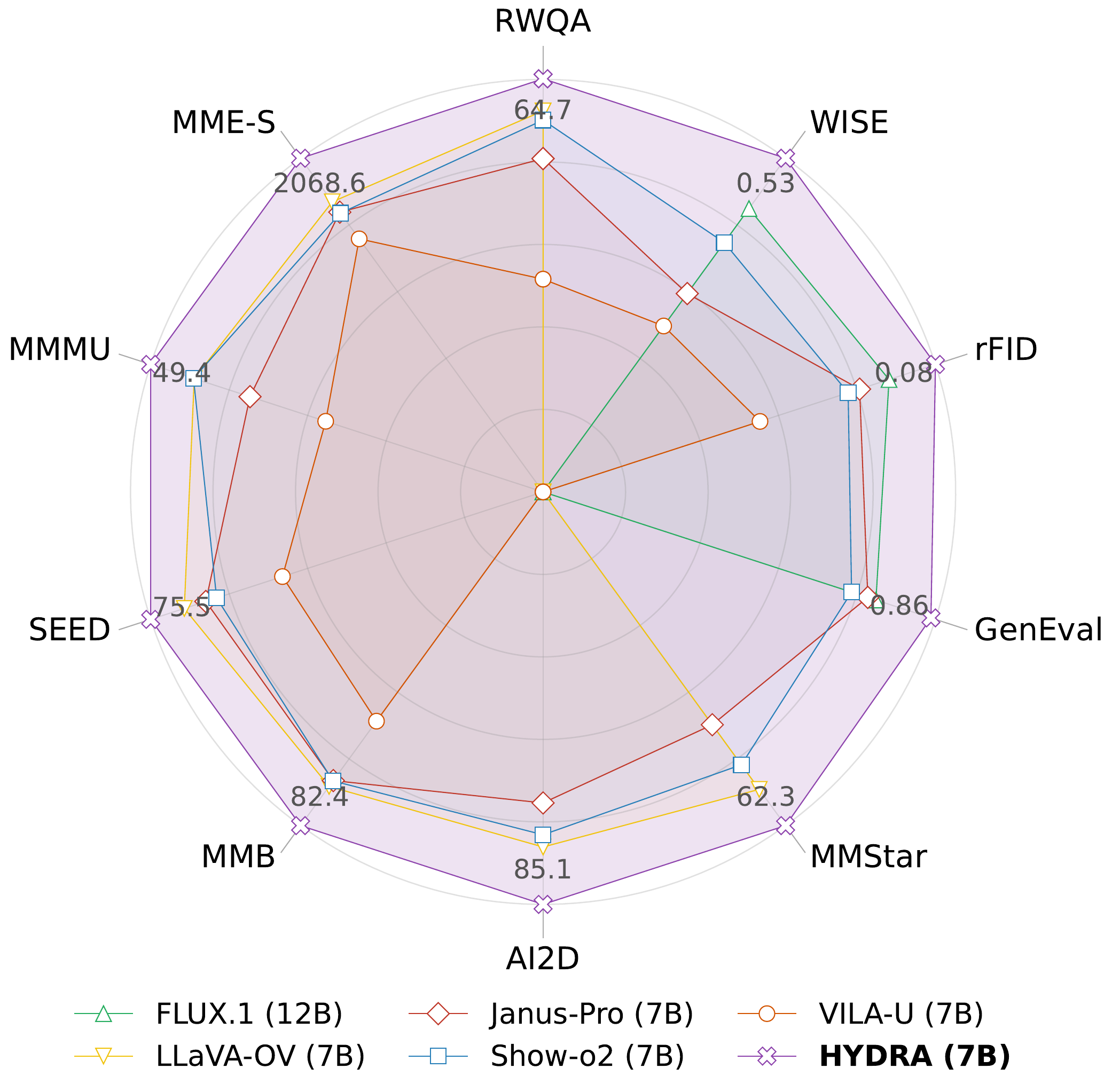}}
    \hfill
    \subcaptionbox{Image generation results.}
        {\includegraphics[width=0.6\linewidth]{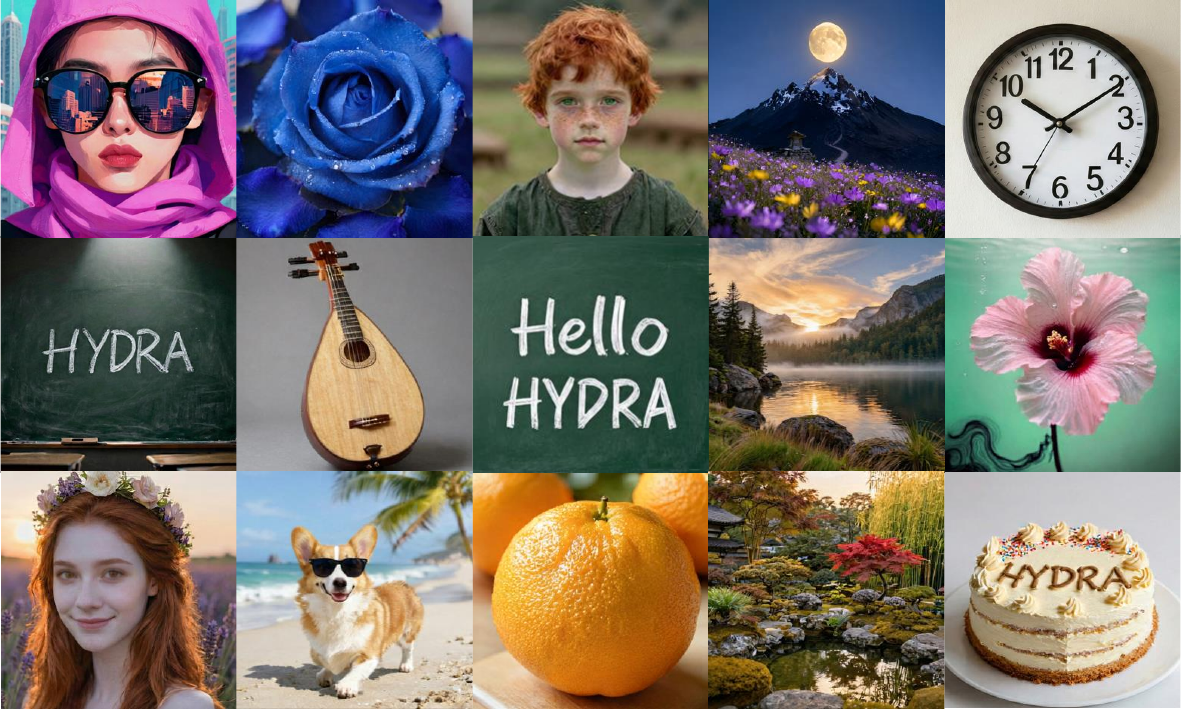}}
    \vspace{-1mm}
    
    \caption{\textbf{Multimodal understanding and image generation results.} 
    (a) \model outperforms unified and task-specific models across diverse benchmarks.
(b) \model generates high-fidelity images with strong semantic alignment.} 
    \label{fig:full_figure}
\end{center}
\end{figure}
\vspace{-3mm}
\begin{abstract}
Unified Multimodal Models struggle to bridge the fundamental gap between the abstract representations needed for visual understanding and the detailed primitives required for generation.  Existing approaches typically compromise by employing decoupled encoders, stacking representation encoder atop VAEs, or utilizing discrete quantization. However, these methods often disrupt information coherence and lead to optimization conflicts. 
To this end, we introduce \tokenizer, a representation-harmonized pure ViT in the insight that visual modeling should evolve from generation to understanding. \tokenizer reformulates the standard backbone into a progressive learner that transitions from a Gen-ViT, which captures structure-preserving primitives, to a Sem-ViT for semantic encoding. Crucially, this transition is mediated by a Generation-Semantic Bottleneck (GSB), which compresses features into a low-dimensional space to filter noise for robust synthesis, then restores dimensionality to empower complex semantic comprehension. Built upon this foundation, we present \model, a native unified framework integrating perception and generation within a single parameter space. 
Extensive experiments establish \model as a new state-of-the-art. It sets a benchmark in visual reconstruction (rFID 0.08) and achieves top-tier generation performance on GenEval (0.86), DPG-Bench (86.4), and WISE (0.53), while simultaneously outperforming previous native UMMs by an average of $\sim$10.0 points across eight challenging understanding benchmarks. 

\end{abstract}


\begin{figure*}[!t]
     \centering
     \includegraphics[width=0.95\linewidth]{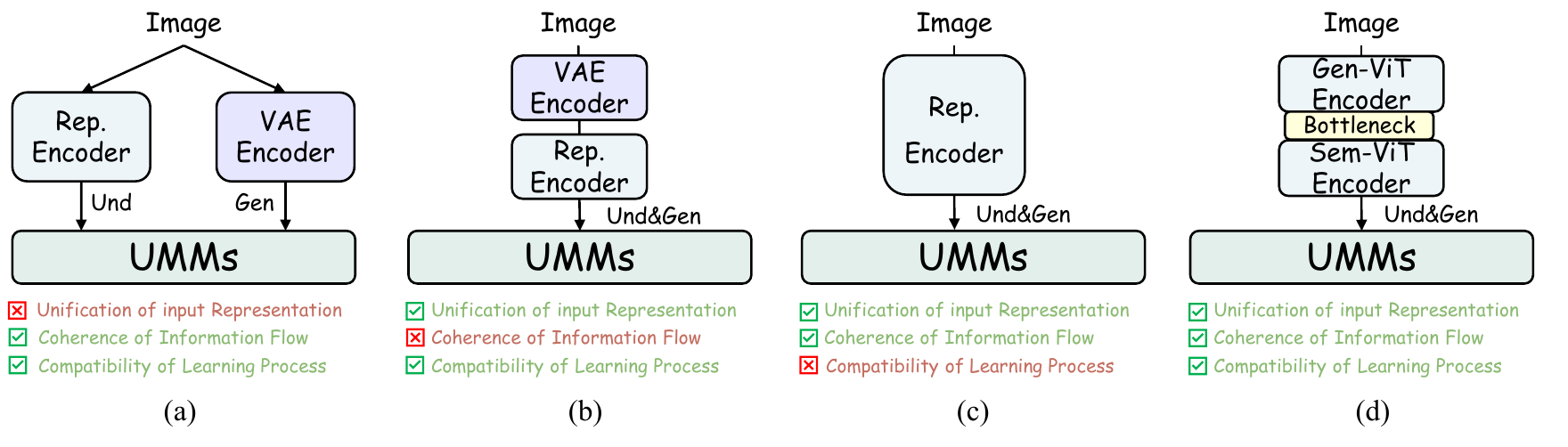}
     \caption{\textbf{Representation schemes in native unified multimodal models.} (a) Decoupled Encoder \cite{bagel,wu2025janus}: Uses separate encoders (a VAE and a representation encoder) for generation and understanding. (b) Sequential Encoder \cite{xie2025show}: Feeds the outputs of the VAE directly into the representation encoder in a cascaded manner. (c) Single Encoder \cite{unitok,wu2025harmonizing}: Unifies both tasks within a standalone representation encoder. (d) Our Proposed Representation-Harmonized ViT Design: Employs a single ViT backbone with a bottleneck module to harmonize feature learning across both tasks.}
     \vspace{-4mm}
     \label{fig:intro}
\end{figure*}
\section{Introduction}
\label{sec:intro}
\vspace{-2mm}
Unifying visual understanding and generation has emerged as a pivotal frontier in multimodal intelligence~\cite{bagel, cao2025hunyuanimage, mogao}. Native unified multimodal models (UMMs)~\cite{wu2025harmonizing, zhou2024transfusion, xie2025show} advocate for direct decoding within a unified parameter space, demonstrating superior synergy over composite UMMs~\cite{seed-x, unilip, blip3-o}. However, achieving rational unification is hindered by a fundamental representational divergence, stemming from the fact that understanding and generation constitute inverse tasks with conflicting demands: the former necessitates high-level semantic abstractions, whereas the latter requires compact structural primitives for fine-grained synthesis~\cite{clip,vae}. This intrinsic conflict forces existing frameworks into disjointed, asymmetric designs, significantly increasing architectural complexity and optimization difficulty.

We attribute this dilemma primarily to the structural limitations of existing image tokenizers, which fail to simultaneously satisfy the three critical criteria illustrated in Fig.~\ref{fig:intro}.
First, decoupled paradigms that employ separate encoders for understanding and generation ~\cite{bagel,cao2025hunyuanimage} inherently lack \textit{Unification of Input Representation}, relying on disjoint features that sever the synergy between the two tasks.
\qxr{Second, sequential architectures that stack representation encoders atop VAEs~\cite{xie2025show,liu2025tuna} theoretically unify the input but compromise the \textit{Coherence of Information Flow} due to the significant representation mismatch between the generative VAE latent space and the semantic features required by the representation encoder.}
Third, while utilizing a single shared representation encoder~\cite{unitok,wu2025harmonizing,jiao2025unitoken} attempts to solve this, it often suffers from poor \textit{Compatibility of Learning Process}, where the conflicting objectives of high-frequency detail preservation and semantic abstraction lead to optimization difficulties.
Consequently, current methods face an unavoidable trade-off: they either sacrifice generative fidelity, lose semantic alignment, or struggle to converge on a shared representation.

To address this, we propose HYDRA-TOK, a representation-harmonized pure ViT framework. \qxr{Distinct from previous methods that employ convolutional VAEs for visual tokenization, HYDRA-TOK jointly trains reconstruction and understanding capabilities natively within a single ViT backbone.} Our core design principle is built on a key insight: a compact feature space capable of reconstructing input data can serve as a robust foundation for semantic understanding. This reconstruction task functions as an information bottleneck, compelling the compact feature to discard extraneous details and instead acquire a vocabulary of dense, structural primitives. These primitives provide a solid basis, thereby enabling the model to construct semantic abstractions from the ground up.  

Driven by this insight, we reformulate a ViT-based representation encoder \cite{internvl2.5,internvl} into a progressive learner that transitions from a Generation vision transformer (Gen-ViT), which captures structure-preserving primitives for high-fidelity synthesis, to a Semantic vision transformer (Sem-ViT). To unify these distinct objectives within a single model, we introduce the Generation-Semantic Bottleneck (GSB). The GSB is architected around a novel compress-and-reconstruct operation, creating an information bottleneck that fosters both high-level semantic abstraction and detailed generative fidelity. Specifically, GSB first compresses features into a compact low-dimension space to filter out noise component, then reconstructs them to the original space for subsequent semantic distillation. In this manner, the compact features can both encode structured details while maintaining semantic awareness.

Built upon \tokenizer, we present \model, a unified framework that achieves complete architectural and representational unification. Leveraging the coherent visual representations provided by \tokenizer, visual signals are processed as sequences via a dual-head mechanism, employing an autoregressive head for text prediction and a rectified flow matching head for image generation. Extensive experiments confirm that \model achieves superior performance, highlighting the harmony between understanding and generation during joint training. In terms of multimodal understanding, \model outperforms existing native UMMs by an average margin of approximately 10.0 points across eight benchmarks. Meanwhile, it establishes a new benchmark in visual reconstruction with a remarkable rFID of 0.08, providing a robust foundation that facilitates state-of-the-art generation records: 0.86 on GenEval \cite{ghosh2023geneval}, 86.4 on DPG-Bench \cite{hu2024ella}, and 0.53 on WISE \cite{niu2025wise}. Additionally, we find that joint training consistently outperforms separate training for both generation and understanding, validating the effectiveness of our harmonized representation. Our contributions are summarized as follows:
\begin{itemize}
    \item We propose \tokenizer, a {representation-harmonized} pure ViT that resolves the understanding-generation conflict via a progressive learner, unifying input representations without quantization errors.
    
    \item We present \model, a native unified framework that integrates understanding and generation within a single parameter space, utilizing a dual-head mechanism for seamless task execution. 
    \item Empirical results demonstrate that \model outperforms native UMMs by $\sim$10.0 points on understanding benchmarks and achieves state-of-the-art performance on GenEval, DPG-Bench, and WISE.
\end{itemize}

\section{Method}
\label{sec:method}
\begin{figure*}[!t]
    \centering
    \includegraphics[width=0.95\linewidth]{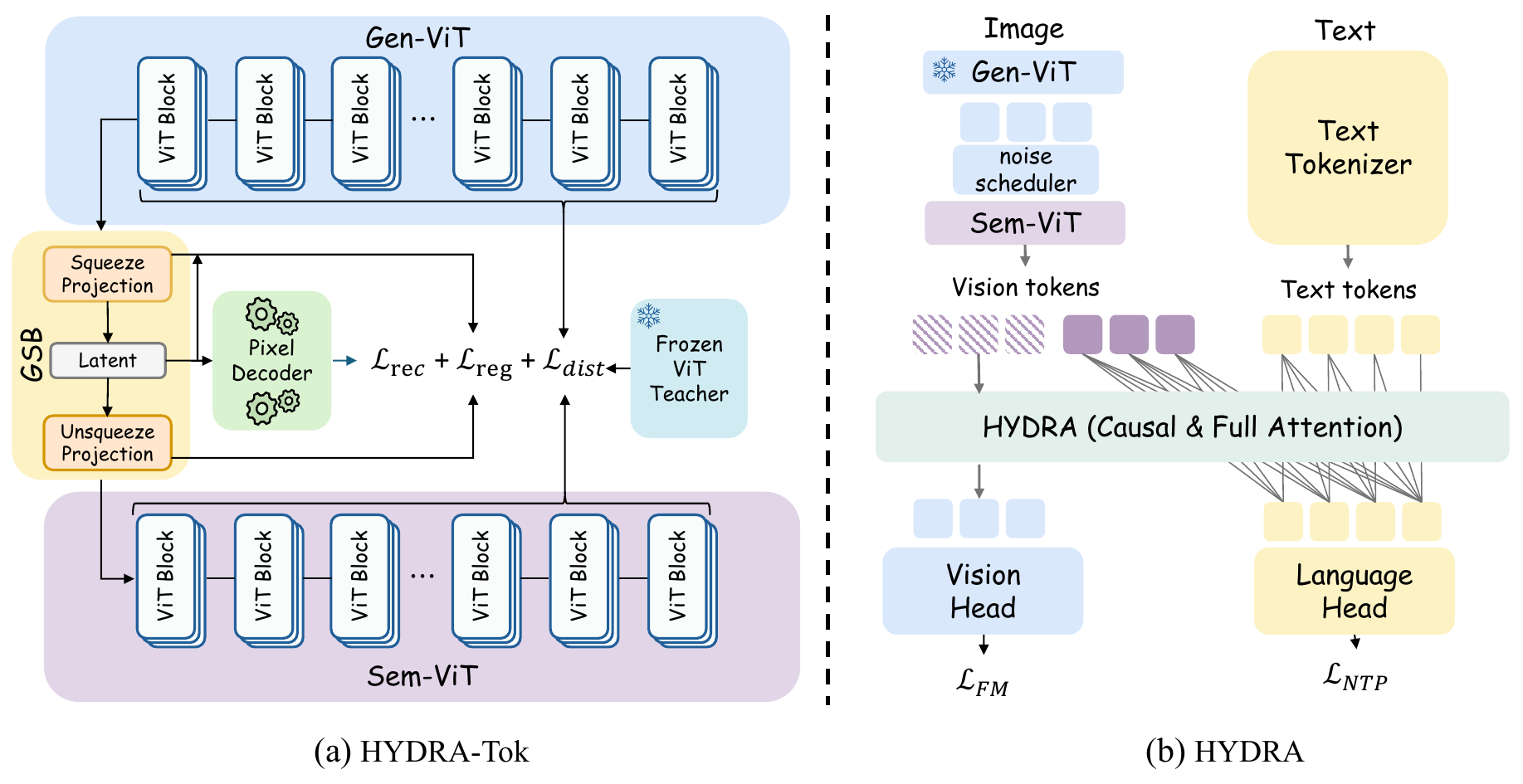}
    \vspace{-2mm}
    \caption{\textbf{Training process illustration for \tokenizer and \model.} (a) \tokenizer functions as a progressive learner, bridging the gap between reconstruction and understanding. It employs a Generation-Semantic Bottleneck (GSB) to execute a unique compress-and-reconstruct operation, effectively filtering noise to transition from structure-preserving primitives (Gen-ViT) to semantic abstractions (Sem-ViT). (b) \model unifies these representations using a dual-head architecture, effectively combining autoregressive text prediction and rectified flow matching.}
    \label{fig:main}
   
\end{figure*}

\vspace{-2mm}
We present \model, a unified framework harmonizing visual understanding and generation. Central to our approach is \tokenizer, a pure ViT grounded in a key insight: a compact feature space capable of reconstructing inputs serves as a robust foundation for semantic understanding. Adopting a functionally progressive learner design (Fig.~\ref{fig:main}), we transition continuously from structure-preserving primitives (Gen-ViT) to semantic abstractions (Sem-ViT). This is realized via the Generation-Semantic Bottleneck (GSB) and its novel compress-and-reconstruct operation. By compressing features to filter noise and reconstructing them for distillation, the GSB effectively balances generative fidelity with semantic awareness.

\subsection{\tokenizer}
\label{subsec:tokenizer}

Traditional tokenizers face a rigid trade-off between preserving semantic depth and maintaining structural detail. To resolve this issue, \tokenizer reformulates the complete vision transformer backbone into three functionally distinct yet continuous components, effectively followed by a lightweight flow-based decoder.

\paragraph{Generation Vision Transformer (Gen-ViT).}
Given an input image $\mathbf{x} \in \mathbb{R}^{H \times W \times 3}$, we first flatten and project non-overlapping patches into continuous embeddings $\mathbf{H}_0 \in \mathbb{R}^{N \times D}$, where $N$ and $D$ correspond to the number of tokens and dimension. The initial stage, Gen-ViT, is tasked with extracting low-level structural primitives essential for generation. Unlike standard encoders that aggressively compress spatial information, Gen-ViT preserves fine-grained spatial covariance:
\begin{equation}
    \mathbf{H}_{\mathtt{mid}} = \text{Gen-ViT}(\mathbf{H}_0) = \Phi_{L_{\mathtt{gen}}} \circ \dots \circ \Phi_{1}(\mathbf{H}_0),
\end{equation}
where $\Phi_l$ denotes the $l$-th transformer block. This phase ensures that the latent space retains the structural foundation required for high-fidelity synthesis.

\paragraph{Generation-Semantic Bottleneck (GSB).}
To transition from structure-preserving primitives to semantic abstractions, we introduce the GSB block. Built on a  compress-and-reconstruct operation, GSB serves as an information bottleneck that filters extraneous noise to balance the conflicting demands of understanding and generation. As evidenced by our ablation (Fig.~\ref{fig:channel_ablation_combined}), while higher dimensions ($C$) aid reconstruction and understanding, they cause generation performance to collapse (e.g., at $C \ge 256$). This confirms that excessive dimensionality introduces redundancy that disrupts generative stability.

To resolve this, GSB acts as a stabilization pivot by first {compressing} the intermediate features $\mathbf{H}_{\mathtt{mid}}$ into a compact probabilistic space via a lightweight projector $\mathbf{W}_{\mathtt{proj}} \in \mathbb{R}^{D \times 2C}$, where $C \ll D$ (typically 64):
\begin{equation}
    [\boldsymbol{\mu}, \boldsymbol{\rho}] = \mathbf{H}_{\mathtt{mid}}\mathbf{W}_{\mathtt{proj}}, \quad \mathbf{z} = \boldsymbol{\mu} + \boldsymbol{\epsilon} \odot \exp(0.5 \boldsymbol{\rho}),
    \label{GSB}
\end{equation}
where $\boldsymbol{\mu}, \boldsymbol{\rho} \in \mathbb{R}^{N \times C}$ represent the mean and log-variance, and $\boldsymbol{\epsilon} \sim \mathcal{N}(\mathbf{0}, \mathbf{I})$ is reparameterization noise. 

To structure this latent space, we impose a KL divergence loss that aligns the posterior with a standard normal prior:
\begin{equation}
    \mathcal{L}_{\mathtt{KL}} = -\frac{1}{2} \sum_{j=1}^{C} \left( 1 + \boldsymbol{\rho}_j - \boldsymbol{\mu}_j^2 - \exp(\boldsymbol{\rho}_j) \right).
\end{equation}
To maintain a coherent flow of information under compression and provide a sufficient foundation for subsequent semantic extraction, we introduce a consistency loss $\mathcal{L}_{\mathtt{cos}}$. This loss forces the unprojected features $\mathbf{H}_{\mathtt{bn}} = \boldsymbol{\mu}\mathbf{W}^{\mathtt{und}}_{\mathtt{unproj}}$ to maintain directional alignment with the pre-bottleneck features $\mathbf{H}_{\mathtt{mid}}$:
\begin{equation}
    \mathcal{L}_{\mathtt{cos}} = 1 - \frac{\mathbf{H}_{\mathtt{mid}} \cdot \mathbf{H}_{\mathtt{bn}}}{\| \mathbf{H}_{\mathtt{mid}} \|_2 \| \mathbf{H}_{\mathtt{bn}} \|_2}.
\end{equation}
Based on our experiments, we set the regularization weights to $\lambda_{\mathtt{KL}}=10^{-4}$ and $\lambda_{\mathtt{cos}}=1.0$. The composite bottleneck objective is defined as $\mathcal{L}_{\mathtt{reg}} = \lambda_{\mathtt{KL}}\mathcal{L}_{\mathtt{KL}} + \lambda_{\mathtt{cos}}\mathcal{L}_{\mathtt{cos}}$.

\paragraph{Semantic Vision Transformer (Sem-ViT).}
As the final stage of our functionally progressive learner, Sem-ViT acts as a deep non-linear mapper. Its primary role is to transition the structural foundations (encoded in $\mathbf{H}_{\mathtt{bn}}$) into a high-dimensional semantic space, thereby achieving rational disentanglement:
\begin{equation}
    \mathbf{H}_{\mathtt{out}} = \text{Sem-ViT}(\mathbf{H}_{\mathtt{bn}}) = \Phi_{L_{\mathtt{total}}} \circ \dots \circ \Phi_{L_{\mathtt{gen}}+1}(\mathbf{H}_{\mathtt{bn}}).
\end{equation}
To ensure robust representation learning across the entire hierarchy, we employ semantic self-distillation on {both Gen-ViT and Sem-ViT}. We align the intermediate features from distinct depths of the student with a frozen, pre-trained ViT \cite{internvl2.5} via cosine similarity maximization:
\begin{equation}
    \mathcal{L}_{\mathtt{dist}} = \sum_{l \in \mathcal{S}_{\mathtt{gen}} \cup \mathcal{S}_{\mathtt{sem}}} \left( 1 - \frac{\mathbf{H}^{(l)}(\mathbf{x}) \cdot \mathcal{T}^{(l)}(\mathbf{x})}{\| \mathbf{H}^{(l)}(\mathbf{x}) \|_2 \| \mathcal{T}^{(l)}(\mathbf{x}) \|_2} \right),
\end{equation}
where $\mathcal{S}_{\mathtt{gen}}$ and $\mathcal{S}_{\mathtt{sem}}$ denote the selected layers from Gen-ViT and Sem-ViT, respectively.

\paragraph{Pixel Flow Decoder.}
To unburden the backbone, we employ a lightweight decoder $\mathbf{v}_\theta$ that utilizes flow matching to recover high-frequency details. Conditioned on the sampled latent $\mathbf{c} = \mathbf{z}$ from the GSB (Eq.~\ref{GSB}), it learns to regress the velocity field by minimizing:
\begin{equation}
    \mathcal{L}_{\mathtt{FM}} = \mathbb{E}_{t, \mathbf{x}, \boldsymbol{\epsilon}} \left[ \| \mathbf{v}_\theta(\mathbf{x}_t, t, \mathbf{c}) - (\boldsymbol{\epsilon} - \mathbf{x}) \|^2 \right].
\end{equation}
To further enhance perceptual fidelity, we enforce an LPIPS loss $\mathcal{L}_{\mathtt{lpips}}$ on the estimated clean image $\hat{\mathbf{x}}$ and incorporate an adversarial GAN loss $\mathcal{L}_{\mathtt{gan}}$ to refine texture realism. We empirically set $\lambda_{\mathtt{FM}}=1.0$, $\lambda_{\mathtt{perc}}=0.1$, and $\lambda_{\mathtt{gan}}=0.075$. The total reconstruction loss is formulated as $\mathcal{L}_{\mathtt{rec}} = \lambda_{\mathtt{FM}}\mathcal{L}_{\mathtt{FM}} + \lambda_{\mathtt{perc}}\mathcal{L}_{\mathtt{lpips}} + \lambda_{\mathtt{gan}}\mathcal{L}_{\mathtt{gan}}$.

\paragraph{Total objective.}
Finally, the unified tokenizer is optimized by minimizing the weighted sum of the reconstruction, regularization, and alignment objectives defined above:
\begin{equation}
    \mathcal{L}_{\mathtt{tokenizer}} = \mathcal{L}_{\mathtt{rec}} + \mathcal{L}_{\mathtt{reg}} + \lambda_{\mathtt{dist}}\mathcal{L}_{\mathtt{dist}},
\end{equation}
where the distillation weight is set to $\lambda_{\mathtt{dist}} = 1.0$ by default.

\subsection{\model}
\vspace{-2mm}
\label{subsec:unified_model}

Built upon the robust representations of \tokenizer, \model represents a {native unified framework} that integrates understanding and generation within a single parameter space. By leveraging the coherent visual representations from our tokenizer, \model processes visual and textual signals as a unified sequence via a shared autoregressive transformer, employing a specialized dual-head mechanism to reconcile their distinct output modalities.

\paragraph{Unified Input Representation.}
To achieve the \textit{Unification of Input Representation}, we integrate visual signals into the LLM by treating them as continuous sequences fully compatible with the embedding space. For a given image, we extract its latent representation $\mathbf{H}_{\mathtt{bn}}$ from the GSB. We implement a task-aware injection strategy: for flow-based generation, we introduce diffusion noise corresponding to the time-step $t$ ($\mathbf{H}_{\mathtt{in}} = \mathbf{H}_{\mathtt{bn}} + t\boldsymbol{\epsilon}$); conversely, for understanding, we utilize the clean, deterministic latents to maximize semantic clarity. These processed latents are mapped to the LLM's dimension via a linear projector to yield $\mathbf{U}_{\mathtt{vis}}$. Finally, we concatenate them with text embeddings $\mathbf{E}_{\mathtt{text}}$ to form a unified input stream:
\begin{equation}
    \mathbf{U}_{\mathtt{in}} = [\mathbf{E}_{\mathtt{text}}, \mathbf{U}_{\mathtt{vis}}].
\end{equation}

\paragraph{Dual-Head Decoding.}
The unified sequence $\mathbf{U}_{\mathtt{in}}$ is processed by the shared LLM backbone. Following standard UMM practices \cite{xie2025show,bagel}, we apply a causal attention mask on language tokens and a bidirectional attention mask on visual tokens within the LLM decoder layers. To address the \textit{Compatibility of Learning Process} between discrete text prediction and continuous image synthesis, we branch the output of the final transformer layer into two specialized heads. The Language Head (autoregressive) functions as a standard linear classifier, predicting the probability distribution over the text vocabulary based on the unified context: $\mathcal{L}_{\mathtt{NTP}} = -\sum \log P(\mathbf{y}_i | \mathbf{y}_{<i}, \mathbf{U}_{\mathtt{in}})$. Simultaneously, the Vision Head (diffusion) serves as a continuous regression head modulated by time-step embeddings via AdaLN-Zero~\cite{dit}. This head operates explicitly on the hidden states corresponding to visual tokens ($\mathbf{H}_{\mathtt{LLM}}^{\mathtt{vis}}$) to predict the flow velocity:
\begin{equation}
    \mathbf{v}_{\mathtt{pred}} = \mathtt{Head}_{\mathtt{flow}}(\mathtt{AdaLN}(\mathbf{H}_{\mathtt{LLM}}^{\mathtt{vis}}, t_{\mathtt{emb}})).
\end{equation}
\vspace{-2mm}
\paragraph{Unified Training Objective.}
\model is trained end-to-end via a composite objective:
\begin{equation}
    \mathcal{L}_{\mathtt{total}} = \mathcal{L}_{\mathtt{NTP}} + \mathcal{L}_{\mathtt{FM}}.
\end{equation}
By sharing the LLM backbone while specializing only the input and output layers, \model achieves rational unification.

\subsection{Training Recipe}
To cultivate the {harmonized nature} of \model, we implement a three-stage progressive training strategy. We initialize \tokenizer via reconstruction on large-scale corpora combined with semantic self-distillation against a teacher ViT \cite{internvl2.5}. Subsequent stages incrementally refine the compatibility between understanding and generation (see Appendix \ref{training:details} for full data details).
\vspace{-2mm}
\paragraph{Stage I: Unified Representation Alignment.} 
To resolve the representation divergence at the input level, we freeze the LLM and exclusively tune the vision components. Utilizing 100M image-text pairs, this phase aligns the visual latent space with the linguistic domain, ensuring a coherent unified input representation.
\vspace{-2mm}
\paragraph{Stage II: Comprehensive Multimodal Pre-training.} 
We unlock all parameters to facilitate {harmonized co-promotion} within a single unified stream. The model is jointly optimized on a balanced mix of 30M understanding samples and 30M generative samples ({strategically filtered from Stage I}). This full-parameter update ensures the \textit{Compatibility of Learning Process}, allowing the two tasks to mutually reinforce each other.
\vspace{-2mm}
\paragraph{Stage III: High-Quality Instruction Fine-tuning.} 
The final stage focuses on high-fidelity refinement using curated datasets. We employ 3.2M balanced instruction-tuning samples for understanding, alongside 10M aesthetic-filtered images ({derived from Stage II}) and 6M high-fidelity synthetic images for generation.

\begin{table*}[!t]
\centering
\caption{\textbf{Evaluation on multimodal understanding benchmarks.} \# Params. and \# Data denote the model size and the volume of image-text pairs used for instruction tuning, respectively. Rows in gray indicate  models exceeding 14B parameters.
}
\vspace{-2mm}
\begin{adjustbox}{max width=0.95\linewidth} 
\begin{tabular}{lccccccccccc}
\toprule
\multirow{1}{*}{\textbf{Models}} & \multirow{1}{*}{\textbf{ \# Params}} & \multirow{1}{*}{\textbf{ \# Data}} & \textbf{MME-S} & \textbf{AI2D} & \textbf{OCRBench} & \textbf{MMB} & \textbf{RWQA} & \textbf{SEED} & \textbf{MMMU} & \textbf{MMStar} & \textbf{Avg.}\\ 
\midrule
\multicolumn{12}{c}{\textit{Understanding-only Models}} \\
\midrule
Qwen2.5-VL \cite{bai2023qwen} & 3B & 1B+ & 2199.9 & 81.4 & 82.8 & 77.6 & 65.5 & 74.0 & 51.2 & 56.3 & 70.9 \\
Qwen2.5-VL \cite{bai2023qwen} & 7B & 1B+ & 2312.0 & 84.3 & 88.8 & 82.8 & 68.4 & 77.0 & 58.0 & 64.1 & 75.7 \\
LLaVA-1.5 \cite{liu2023visual} & 7B & 0.7M & 1510.7 & 55.5 & 31.8 & 62.3 & 54.8 & 65.8 & 35.7 & 33.1 & 49.1 \\
LLaVA-OV \cite{li2024llava} & 7B & 3.2M & 1998.1 & 81.4 & 62.2 & 80.8 & 66.3 & 76.7 & 48.8 & 61.7 & 68.7 \\ 

\midrule
\multicolumn{12}{c}{\textit{Composite Unified Multimodal Models}} \\
\midrule
BLIP3-o \cite{blip3-o} & 4B & N/A & 2161.2 & - & - & - & 60.4 & 73.8 & 46.6 & - & - \\
{TokenFlow-XL} \cite{qu2025tokenflow} & {14B} & {N/A} & {1922.1} & {-} & {-} & {68.9} & {56.6} & {72.6} & {43.2} & {-} & {-} \\ 
{SEED-X~\cite{seed-x}} & {17B} & {N/A} & {1457.0} & {70.0} & {-} & {70.1} & {49.1} & {65.6} & {35.6} & {-} & {-} \\
\midrule
\multicolumn{12}{c}{\textit{Smaller Native Unified Multimodal Models}} \\
\midrule
Show-o \cite{xie2024show} & 1.3B & 0.7M & 1345.1 & - & - & - & - & 51.5 & 27.4 & - & - \\
JanusFlow \cite{ma2025janusflow} & 1.3B & 1.4M & 1560.1 & 54.2 & - & 69.4 & 41.2 & 70.5 & 29.3 & 40.6 & - \\
Janus-Pro \cite{chen2025janus} & 1.5B & 10M & 1875.1 & 64.5 & - & 75.5 & 52.6 & 68.3 & 36.3 & 43.1 & - \\
Harmon \cite{wu2025harmonizing} & 1.5B & 3.2M & 1476.2 & 57.0 & 11.2 & 67.1 & 49.8 & 67.1 & 38.9 & 35.3 & 47.4 \\
Show-o2 \cite{xie2025show} & 1.5B & 3.2M & 1742.3 & 69.0 & 24.5 & 67.4 & 56.5 & 65.6 & 37.1 & 43.4 & 53.2 \\
\rowcolor{icmlblue} 
\model & 1.5B & 3.2M & \textbf{1877.9} & \textbf{79.4} & \textbf{50.8} & \textbf{76.8} & \textbf{60.7} & \textbf{74.5} & \textbf{39.1} & \textbf{57.0} & \textbf{63.2} \\ 
\midrule
\multicolumn{12}{c}{\textit{Larger Native Unified Multimodal Models}} \\
\midrule
\g{Ming-UniVision~\cite{huang2025ming}} & \g{16B} & \g{20M+} & \g{2023.0} & \g{82.8} & \g{72.4} & \g{78.5} & \g{-} & \g{-} & \g{40.3} & \g{63.7} & \g{-} \\
\g{BAGEL~\cite{bagel}} & \g{14B} & \g{15.2M} & \g{2388.0} & \g{89.2} & \g{73.3} & \g{85.0} & \g{72.8} & \g{78.5} & \g{55.3} & \g{-} & \g{-} \\
MUSE-VL \cite{muse-vl} & 7B & 10M & - & 69.8 & - & 72.1 & - & 69.1 & 39.7 & 49.6 & - \\
Janus-Pro \cite{chen2025janus} & 7B & 10M & 1927.1 & 71.3 & - & 79.2 & 58.0 & 72.1 & 41.0 & 48.3 & - \\
VILA-U \cite{wu2024vila} & 7B & 4M & 1845.8 & - & - & 66.6 & - & 59.0 & 32.2 & - & - \\
Show-o2 \cite{xie2025show} & 7B & 3.2M & 1920.5 & 78.6 & 32.4 & 79.3 & 64.7 & 69.8 & 48.9 & 56.6 & 62.4 \\
\rowcolor{icmlblue} 
\model & 7B & 3.2M & \textbf{2068.6} & \textbf{85.1} & \textbf{57.7} & \textbf{82.4} & \textbf{64.7} & \textbf{75.5} & \textbf{49.4} & \textbf{62.3} & \textbf{68.9} \\ 
\bottomrule
\end{tabular}
\end{adjustbox}
\label{tab:img_und}
\end{table*}

\vspace{-4mm}
\section{Experiments}
\label{sec:exp}
\vspace{-3mm}
We conduct extensive experiments to evaluate how \model addresses the fundamental challenges in unified multimodal learning identified in the Introduction:
\begin{itemize}
    \item \textbf{RQ1}: What is the sweet spot between  high-level semantic abstraction and compact structural synthesis in \tokenizer to achieve a rational unified representation?
    
    \item \textbf{RQ2}: Can a pure ViT-based tokenizer \tokenizer achieve state-of-the-art reconstruction fidelity?
    
    \item \textbf{RQ3}: Does the pure ViT architecture \model preserve the coherence of information flow better than VAE-based or decoupled approaches, thereby enhancing multimodal understanding?
    
    \item \textbf{RQ4}: Does the joint optimization of autoregressive and flow-matching objectives achieve compatibility, enabling state-of-the-art generative performance?
\end{itemize}

\paragraph{Implementation Details.} We evaluate \model at two scales: 1.5B (based on Qwen2.5-1.5B-Instruct~\cite{qwen2.5}) and 7B (Qwen2.5-7B-Instruct). The \tokenizer is initialized with InternViT-2.5~\cite{internvl2.5} to leverage robust visual priors. The training recipe employs a three-stage progressive pipeline optimized with AdamW~\cite{loshchilov2017decoupled}. The specific learning rate schedules for each stage are detailed in Tab.~\ref{tab:training_details}. For comprehensive experimental details and evaluation metrics regarding all ablation studies, please refer to Appendix \ref{subsec:training_data_abla}.

\begin{table*}[t]
\centering
\caption{\textbf{Comprehensive image generation results on {GenEval} \cite{ghosh2023geneval}, {DPG-Bench} \cite{hu2024ella}, and {WISE} \cite{niu2025wise}.} \# Data. indicates the number of image-text pairs used for visual generation. $^\dagger$ refers to fintuning with GPT-4o distilled synthetic dataset \cite{blip3-o}. More Visualization can be founded in Appendix \ref{vis}.}
\vspace{-2mm}
\label{tab:combined_generation}
\begin{adjustbox}{max width=0.95\linewidth}
\begin{tabular}{@{}lc c ccc c ccc c ccc @{}}
\toprule
\multirow{2}{*}{\textbf{Models}} & \multirow{2}{*}{\textbf{\# Params}} & \multirow{2}{*}{\textbf{\# Data}} & \multicolumn{3}{c}{\textbf{GenEval}} & & \multicolumn{3}{c}{\textbf{DPG-Bench}} & & \multicolumn{3}{c}{\textbf{WISE}} \\
\cmidrule(lr){4-6} \cmidrule(lr){8-10} \cmidrule(lr){12-14}
 & & & \textbf{Two Obj.} & \textbf{Pos.} & \textbf{Over.} & & \textbf{Glob.} & \textbf{Rel.} & \textbf{Over.} & & \textbf{Cult.} & \textbf{Space} & \textbf{Over.} \\ \midrule

\multicolumn{14}{c}{\textit{Generation-only Models}} \\
\midrule
SD3-Med \cite{esser2024scaling} & 2B & 1B+ & 0.94 & 0.33 & 0.74 & & - & - & 84.08 & & - & - & - \\
FLUX.1 [Dev] \cite{flux} & 12B & 1B+ & 0.93 & 0.68 & 0.82 & & 82.10 & 91.10 & 84.00 & & 0.48 & 0.62 & 0.50 \\ \midrule
\multicolumn{14}{c}{\textit{Composite Unified Multimodal Models}} \\
\midrule
Blip3-o$^{\dagger}$ \cite{blip3-o} & 4B & 30M &- & - & 0.84 & & - & - & 79.36 & & - & - & 0.50 \\
TokenFlow-XL \cite{qu2025tokenflow} & 14B & 60M & 0.60 & 0.16 & 0.55 & & {87.53} & {90.30} & {85.02} & & - & - & - \\
SEED-X \cite{seed-x} & 17B & 158M+ & 0.58 & 0.19 & 0.49 & & - & - & - & & - & - & - \\ \midrule
\multicolumn{14}{c}{\textit{Smaller Native Unified Multimodal Models}} \\
\midrule

Show-o \cite{xie2024show} & 1.3B & 36M & 0.52 & 0.11 & 0.53 & & - & - & 67.27 & & - & - & 0.35 \\
Harmon \cite{wu2025harmonizing} &1.5B & 100M & 0.86 & 0.74 & 0.76 & & - & - & - & & 0.38 & 0.52 & 0.41 \\
Show-o2 \cite{xie2025show} & 1.5B & 66M & 0.86 & 0.46 & 0.73 & & {87.53} & {90.30} & {85.02} & & 0.33 & 0.53 & 0.39 \\
\rowcolor{icmlblue}
\model & 1.5B & 100M & \textbf{0.95} & \textbf{0.76} & \textbf{0.86} & & \textbf{89.65} & \textbf{91.59} & \textbf{85.51}& & \textbf{0.49} & \textbf{0.64} & \textbf{0.49} \\
\midrule

\multicolumn{14}{c}{\textit{Larger Native Unified Multimodal Models}} \\
\midrule
\g{Ming-UniVision~\cite{huang2025ming}} & \g{16B} & \g{1B+} & \g{0.93} & \g{0.92} & \g{0.85} & & \g{-} & \g{-} & \g{82.12} & & \g{-} & \g{-} & \g{-} \\
\g{BAGEL~\cite{bagel}} & \g{14B} & \g{1.6B} & \g{0.95} & \g{0.78} & \g{0.88} & & \g{88.94} & \g{90.82} & \g{85.07} & & \g{0.44} & \g{0.68} & \g{0.52} \\
MUSE-VL \cite{muse-vl} & 7B & 24M & 0.64 & 0.25 & 0.57& & - & - & - & & - & - & - \\
Janus-Pro \cite{chen2025janus} & 7B & 144M & 0.89 & 0.79 & 0.80 & & 86.90 & 89.32 & 84.19 & & 0.30 & 0.49 & 0.35 \\
VILA-U \cite{wu2024vila} & 7B & 15M & - & - & - & & -& - & - & & 0.26 & 0.37 & 0.31 \\
Show-o2 \cite{xie2025show} & 7B & 66M & 0.87 & 0.52 & 0.76 & & {89.00} & 91.81 & {86.14} & & 0.40 & 0.58 & 0.44 \\
\rowcolor{icmlblue}
\model & 7B & 100M &\textbf{0.97} & \textbf{0.81} & \textbf{0.86} & & \textbf{90.22} & \textbf{91.83} & \textbf{86.40} & & \textbf{0.52} & \textbf{0.68} & \textbf{0.53} \\
\bottomrule
\end{tabular}
\end{adjustbox}
\end{table*}

\vspace{-4mm}

\begin{figure*}[!t] 
    \centering
    
    \begin{minipage}[b]{0.48\linewidth}
        \centering
        \begin{subfigure}[b]{0.48\linewidth}
            \centering
            \includegraphics[width=\linewidth]{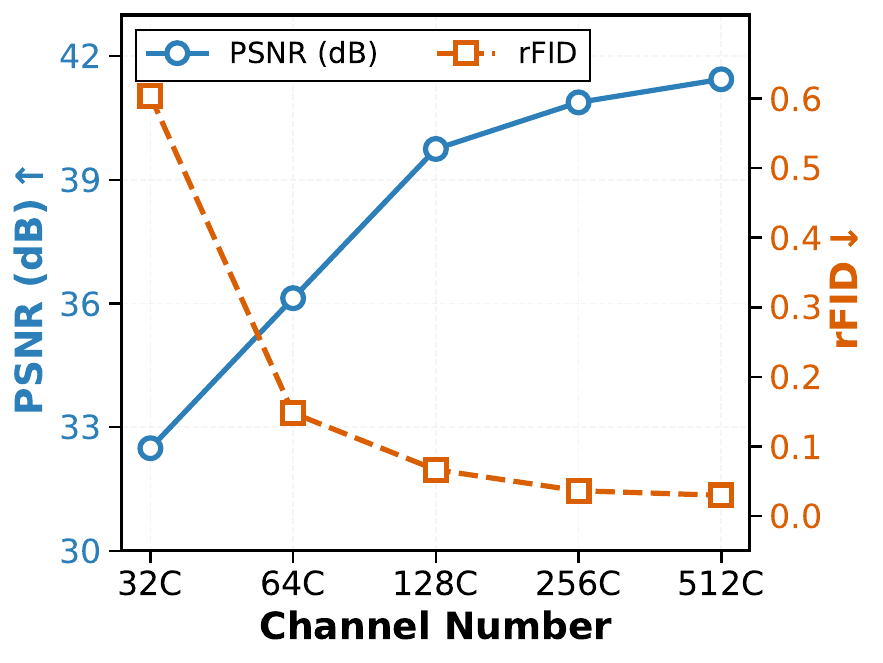}
            \caption{Reconstruction} 
            \label{fig:channel_rec}
        \end{subfigure}
        \hfill
        \begin{subfigure}[b]{0.48\linewidth}
            \centering
            \includegraphics[width=\linewidth]{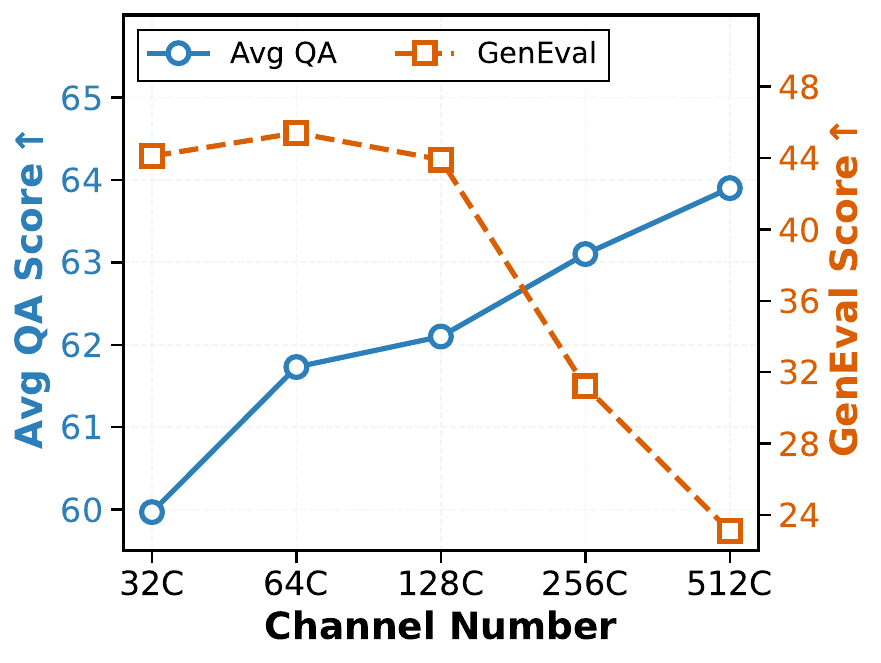}
            \caption{Und. vs. Gen.} 
            \label{fig:channel_und}
        \end{subfigure}
         \vspace{-2mm}
        \caption{\textbf{Ablation results on different latent channel dimensions.}}
        \label{fig:channel_ablation_combined}
    \end{minipage}
    \hfill 
    \begin{minipage}[b]{0.48\linewidth}
        \centering
        \begin{subfigure}[b]{0.48\linewidth}
            \centering
            \includegraphics[width=\linewidth]{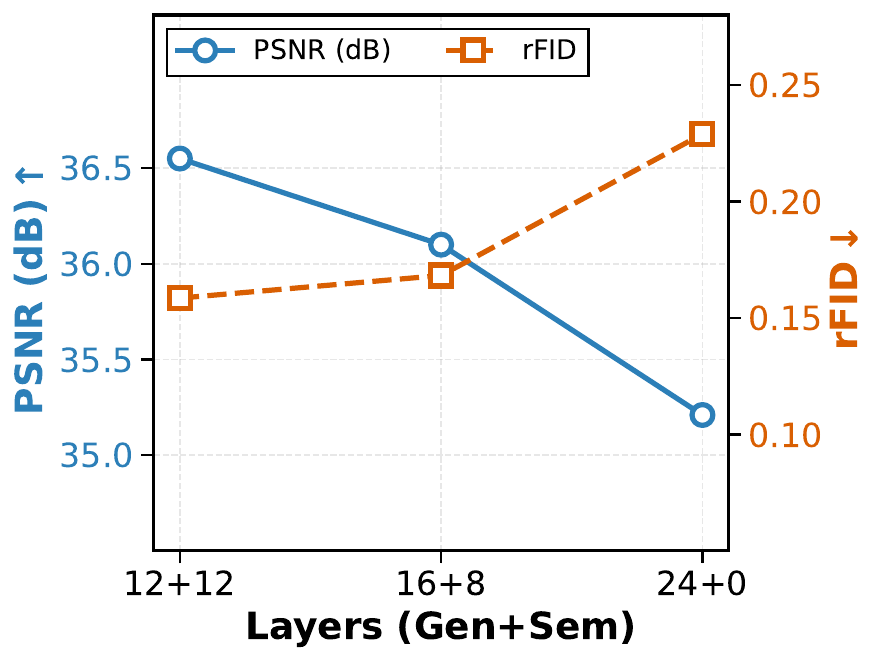}
            \caption{Reconstruction} 
            \label{fig:layer_rec}
        \end{subfigure}
        \hfill 
        \begin{subfigure}[b]{0.48\linewidth}
            \centering
            \includegraphics[width=\linewidth]{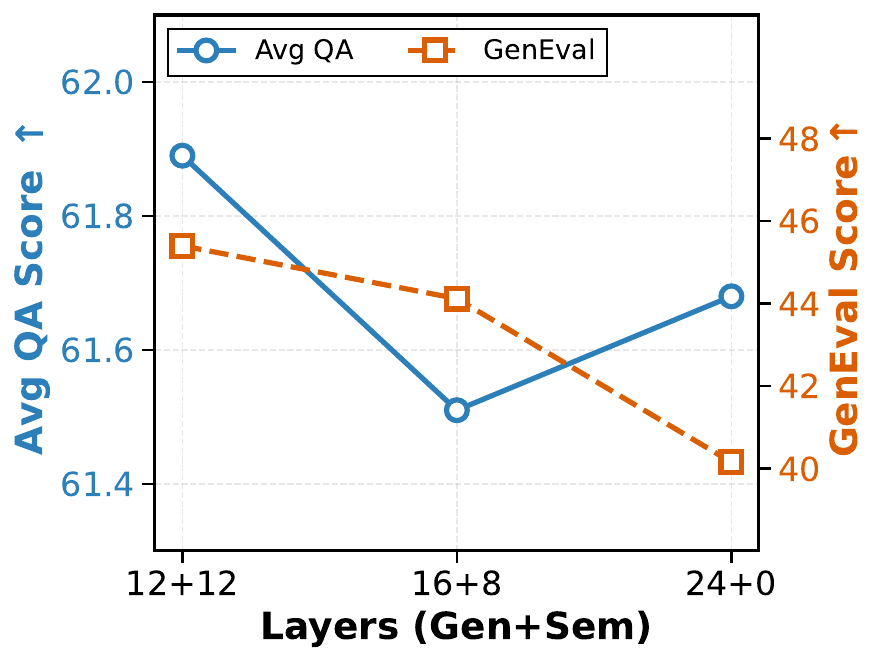}
            \caption{Und. vs. Gen.} 
            \label{fig:layer_und}
        \end{subfigure}
        \vspace{-2mm}
        \caption{\textbf{Ablation results on different layer configurations.}}
        \label{fig:layer_ablation_combined}
    \end{minipage}
    
\end{figure*}
\subsection{Main Results}
\label{subsec:main-results}

\paragraph{RQ1: Identifying the Sweet Spot for Rational Unification.}
Achieving a rational unified representation requires identifying the optimal synergy between high-level semantic abstraction and compact structural synthesis within a single feature space. We seek the architectural ``sweet spot'' where both capabilities are maximized simultaneously, rather than viewing them as competing objectives.
As initially determined regarding channel dimensions (referencing {Fig.~\ref{fig:channel_ablation_combined}}), incorporating the {GSB} to compress the feature space to $C=64$ represents the representational sweet spot, achieving \textit{Unification of Input Representation} by filtering redundant noise while retaining semantic capacity.
Further investigating this balance at the architectural level, Fig. 5 illustrates the impact of different layer allocations. We observe that configurations deviating from a balanced structure fail to achieve optimal synergy. The highly imbalanced $24+0$ configuration (pure Gen-ViT) exhibits the poorest performance across reconstruction fidelity (lowest PSNR, highest rFID), generative capability (lowest GenEval Score), and understanding (Avg QA). Similarly, the $16+8$ configuration also yields suboptimal results across these metrics compared to the balanced approach. Consequently, the $12+12$ configuration clearly emerges as the architectural "sweet spot." It achieves peak performance across all evaluated metrics (highest PSNR, Avg QA, and GenEval Score), effectively harmonizing the depth required for semantic abstraction with the structural stability needed for generation.

\paragraph{RQ2: High-Fidelity Reconstruction.}
We next verify whether our pure ViT-based \tokenizer can serve as a high-quality visual foundation. Historically, ViTs were deemed unsuitable for dense pixel-level synthesis compared to Convolutional VAEs.  On ImageNet-1K under identical $256\times256$ settings ($^\dagger$), \tokenizer achieves 36.39 PSNR and 0.08 rFID, significantly outperforming the unified baseline UniFlow \cite{yue2025uniflow} (33.23 PSNR, 0.26 rFID). At $448\times448$ resolution, it maintains comparable reconstruction quality (31.90 PSNR, 0.12 rFID) against the generative-only FLUX-VAE \cite{flux} (32.91 PSNR, 0.08 rFID). This confirms our \textit{functionally progressive design} enables pure ViTs to successfully supersede traditional VAEs.

\begin{wraptable}{r}{0.45\textwidth} 
    \centering
    \caption{ \textbf{ Reconstruction results on the ImageNet-1K validation set.} $^\dagger$ indicates models trained at a resolution of 448 and resized to 256 for evaluation.
    }
    \label{tab:tokenizer}
    \begin{adjustbox}{max width=\linewidth} 
    \begin{tabular}{l c ccc} 
    \toprule
    \textbf{Method} & \textbf{Res.} & \textbf{PSNR} $\uparrow$ & \textbf{SSIM} $\uparrow$ & \textbf{rFID} $\downarrow$ \\
    \midrule
    \multicolumn{5}{c}{\textbf{\textit{Generative Only Tokenizer}}} \\
    \midrule
    SD-VAE  \cite{sd} & 256 & 23.54 & 0.68 & 1.22 \\
    FLUX-VAE \cite{flux} & 256 & {32.74} & {0.92} & {0.18} \\
    VA-VAE \cite{vavae} & 256 & 27.96 & 0.79 & 0.28 \\
    DC-AE \cite{flowmo} & 512 & 26.15 & 0.71 & 0.22 \\  
     FLUX-VAE \cite{flux} & 448 & {32.91} & {0.92} & {0.08} \\
    \midrule
    \multicolumn{5}{c}{\textbf{\textit{Unified Tokenizer}}} \\
    \midrule
    Tokenflow \cite{qu2025tokenflow} & 384 & 22.73 & 0.73 & 0.63 \\
    UniTok \cite{unitok} & 256 & 27.28 & 0.77 & 0.41 \\
    UniLIP \cite{unilip} & 448 & 24.62 & 0.79 & 0.31\\
    BLIP3-o \cite{blip3-o} & 256 & 14.71 & 0.58 & 3.18 \\
    QLIP \cite{qlip}& 256 & 23.16 & 0.63 & 3.21 \\
    UAE \cite{uae} & 256 & 29.65 & 0.88 & 0.19 \\
    RAE \cite{rae} & 256 & 18.05 & 0.50 & 2.04 \\           
    {UniFlow} $^\dagger$ \cite{yue2025uniflow} & 256 & {33.23} & {0.96} & {0.26} \\
    \rowcolor{icmlblue}
    \textbf{\tokenizer} $^\dagger$ & 256 & \textbf{36.39} & \textbf{0.96} & \textbf{0.08} \\  
    \rowcolor{icmlblue}
    \textbf{\tokenizer} & 448 & \textbf{31.90} & \textbf{0.90} & \textbf{0.12} \\
    \bottomrule 
    \end{tabular}
    \end{adjustbox}
\end{wraptable}
\begin{figure}[!t]
    \centering
    \begin{subfigure}[b]{0.49\linewidth}
        \centering
        \includegraphics[width=\linewidth]{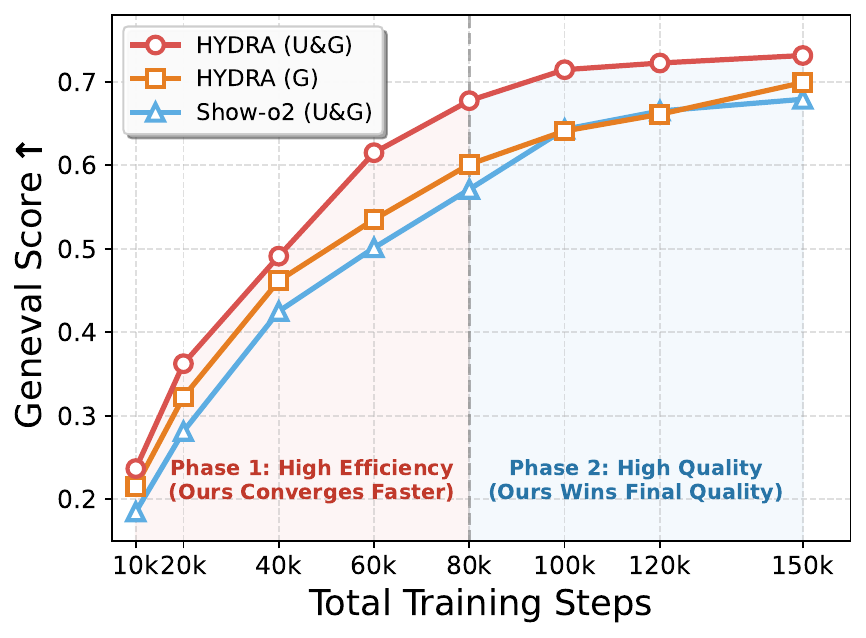}
        \caption{{Generation.} }
        \label{fig:unify_gen}
    \end{subfigure}
    \hfill 
    \begin{subfigure}[b]{0.49\linewidth}
        \centering
        \includegraphics[width=\linewidth]{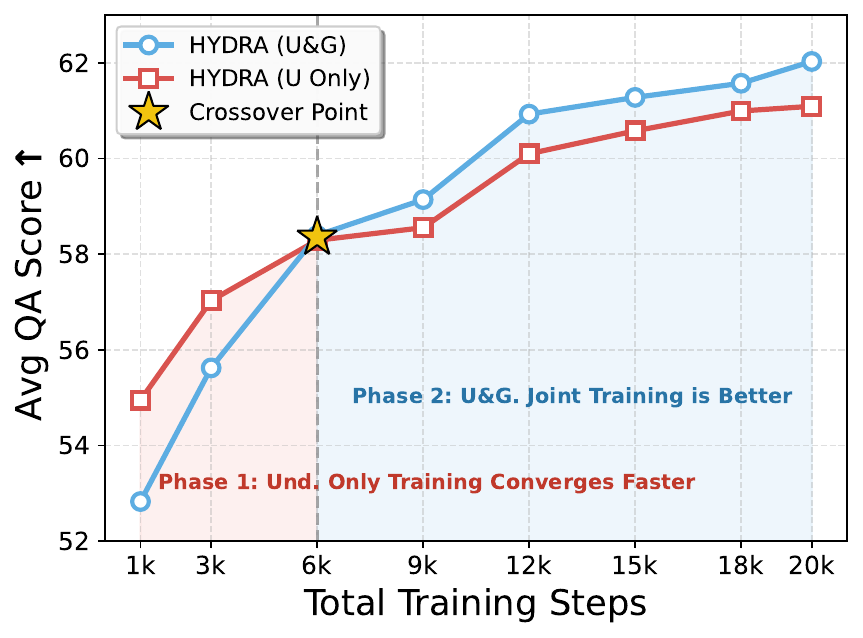}
        \caption{{Understanding.}} 
        \label{fig:unify_und}
    \end{subfigure}
    \caption{\textbf{Analysis of representation-harmonized co-promotion.} (a) For generation, joint training enhances generation efficiency  by stabilizing the latent space. (b) For understanding, joint training surpasses single-task baselines after a crossover point, showing that generative constraints refine perceptual precision. Details of each understanding benchmark and experiment setting  is shown in Fig. \ref{fig:benchmark_comparison} and Appendix \ref{harmon_details}}
    \label{fig:joint}
    \vspace{-4mm}
\end{figure}

\vspace{-3mm}
\paragraph{RQ3: Coherence of Information Flow.} 
We assess the coherence of information flow by evaluating multimodal understanding performance, as shown in Tab.~\ref{tab:img_und}. Sequential architectures, such as Show-o2~\cite{xie2025show}, often compromise information coherence due to the compression bottleneck imposed by  VAEs. In contrast, \model maintains a continuous and uncompressed information flow throughout its pure ViT backbone. At the 1.5B scale, \model demonstrates dominant performance with an average score of {63.1}, surpassing the strong baseline Show-o2 (53.2) by a significant margin. This advantage is particularly pronounced in fine-grained tasks sensitive to information loss, such as OCRBench \cite{liu2024ocrbench}, where \model achieves {50.8}, a score more than double that of Show-o2 (24.5). Scaling up to 7B, \model continues to lead, outperforming Show-o2 on complex reasoning benchmarks like MMStar \cite{chen2024mmstar} ({62.3} vs. 56.6) and SEED \cite{li2023seed} ({75.5} vs. 69.8). Most notably, in OCRBench \cite{liu2024ocrbench}, \model preserves high-frequency character details often discarded by VAE-based unified models, scoring {57.7} compared to 32.4 for Show-o2.

\paragraph{RQ4: Compatibility of Generation Learning Process.}

We begin by investigating the mechanism behind this compatibility through the training dynamics illustrated in Fig.~\ref{fig:joint}. First, regarding generative harmony (Fig.~\ref{fig:unify_gen}), we observe that joint training (U\&G) consistently outperforms single-task generation (G Only). It not only demonstrates higher convergence efficiency in the early stage (Phase 1) but also achieves superior final fidelity (Phase 2). This validates that the semantic alignment provided by understanding tasks effectively stabilizes the generative latent space, leading to faster and better convergence.
Second, regarding understanding harmony (Fig.~\ref{fig:unify_und}), a distinct trend appears. While single-task understanding (U Only) learns faster initially, joint training (U\&G) overtakes it after a critical \textit{crossover point} ($\sim$6k steps). This phenomenon demonstrates that although generation tasks are harder to optimize initially, their fine-grained structural constraints eventually refine the model's perceptual precision, proving that generation and understanding are mutually reinforcing in our framework.
\begin{wrapfigure}{R}{0.48\textwidth}
    \vspace{-10pt} 
    \centering
    \includegraphics[width=\linewidth]{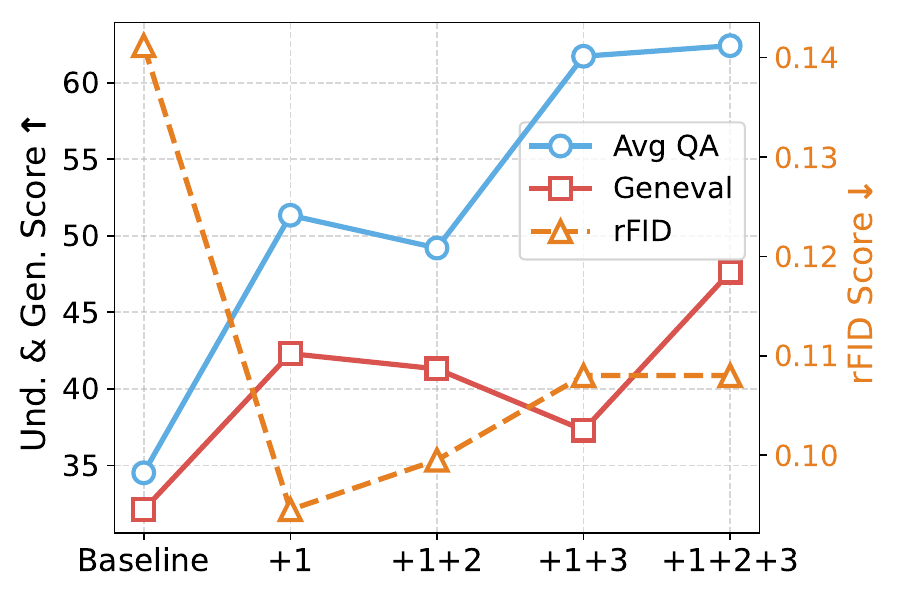}
    \vspace{-5pt}
    \caption{\textbf{Ablation of \tokenizer loss.} Baseline utilizes only $\mathcal{L}_{\text{rec}}$. ``1'': Teacher init; ``2'': $\mathcal{L}_{\text{reg}}$; ``3'': $\mathcal{L}_{\text{dist}}$.}
    \label{fig:tokenizer_loss}
    \vspace{-10pt} 
\end{wrapfigure}

Supported by this harmonized training dynamic, we further validate the learning compatibility on text-to-image generation benchmarks in Tab.~\ref{tab:combined_generation}. A common failure mode in UMMs is the ``tug-of-war'' where improving understanding degrades generation. \model breaks this trade-off, demonstrating that joint optimization leads to superior generative performance. At the 1.5B scale, \model establishes a new benchmark for native UMMs, achieving an Overall GenEval score of {0.86} and DPG-Bench score of {85.51}, significantly outperforming Show-o2 (0.73 / 85.02). At the 7B scale, \model sets new state-of-the-art records with a GenEval Overall score of {0.86}, surpassing both the unified model Ming-UniVision \cite{huang2025ming}  (0.85) and the specialized 12B model FLUX.1 (0.82)~\cite{flux}. Furthermore, \model achieves an overall score of {0.53} on the WISE, demonstrating robust alignment across diverse cultural and spatial contexts compared to existing native unified models.

\subsection{Ablation Analysis}
\paragraph{\tokenizer Training Objectives.}
Fig. \ref{fig:tokenizer_loss} validates our tokenizer's training objectives. The baseline, relying solely on reconstruction loss ($\mathcal{L}_{\text{rec}}$), suffers feature collapse and yields suboptimal results. Teacher initialization provides a crucial semantic scaffold, boosting performance, while distillation ($\mathcal{L}_{\text{dist}}$) further enhances comprehension without compromising structure. Ultimately, the complete objective achieves optimal synergy, balancing understanding, generation, and reconstruction to robustly support the unified architecture.

\begin{wraptable}[12]{r}{0.49\textwidth}
\renewcommand\arraystretch{1.06}
\centering
\setlength{\tabcolsep}{2mm} 
\small
{
\vspace{-12pt}
\caption{
\textbf{Ablation study on the effects of the three training stages.}
}\label{tab:ablation_stage}
\vspace{5pt}
{
\begin{tabular}{ccccc}
\toprule
\textbf{Stage I} & \textbf{Stage II} & \textbf{Stage III} & \textbf{Avg} & \textbf{GenEval} \\
\midrule
\xmark & \cmark & \cmark & 0.719 & 0.82 \\
\cmark & \xmark & \cmark & 0.683 & 0.77 \\
\cmark & \cmark & \xmark & - & 0.84 \\
\rowcolor{icmlblue}
\cmark & \cmark & \cmark & \textbf{0.744} & \textbf{0.86} \\
\bottomrule
\end{tabular}
}
\vspace{-3pt}
}
\end{wraptable}
\paragraph{\model Training Stages.}
Tab. \ref{tab:ablation_stage} confirms the indispensability of each stage in our progressive training recipe. Omitting Stage I causes a universal performance decline, underscoring its role in initial alignment. Removing Stage II severely degrades generation while impairing understanding, proving it critical for consolidating the unified feature space. Skipping Stage III results in a total loss of instruction-following capabilities for QA tasks. Consequently, the full recipe yields peak performance across all benchmarks.


\section{Conclusion}
\label{sec:conclusion}

In this work, we present \tokenizer and \model to reconcile the intrinsic conflict between visual understanding and generation. By employing the Generation-Semantic Bottleneck, \tokenizer functions as a progressive learner that harmonizes structural primitives with semantic abstractions, effectively achieving the \textit{Unification of Input Representation}. This cohesive foundation enables \model to integrate understanding and generation within a single parameter space, satisfying the critical criteria of \textit{Coherence of Information Flow} and \textit{Compatibility of Learning Process}. Our extensive experiments not only establish new state-of-the-art benchmarks but also reveal a fundamental insight: understanding and generation are not competitive objectives but complementary forces that drive mutual enhancement through rational unification. We believe this representation-harmonized paradigm establishes a new standard for native UMMs, offering a promising trajectory toward more versatile and scalable multimodal intelligence.

\section*{Acknowledgements}
This work was supported by the Zhongguancun Academy (Grant No.s 1800302002).



\bibliographystyle{IEEEtran}
\bibliography{reference}

\begin{thebibliography}{10}
\providecommand{\url}[1]{#1}
\csname url@samestyle\endcsname
\providecommand{\newblock}{\relax}
\providecommand{\bibinfo}[2]{#2}
\providecommand{\BIBentrySTDinterwordspacing}{\spaceskip=0pt\relax}
\providecommand{\BIBentryALTinterwordstretchfactor}{4}
\providecommand{\BIBentryALTinterwordspacing}{\spaceskip=\fontdimen2\font plus
\BIBentryALTinterwordstretchfactor\fontdimen3\font minus \fontdimen4\font\relax}
\providecommand{\BIBforeignlanguage}[2]{{%
\expandafter\ifx\csname l@#1\endcsname\relax
\typeout{** WARNING: IEEEtran.bst: No hyphenation pattern has been}%
\typeout{** loaded for the language `#1'. Using the pattern for}%
\typeout{** the default language instead.}%
\else
\language=\csname l@#1\endcsname
\fi
#2}}
\providecommand{\BIBdecl}{\relax}
\BIBdecl

\bibitem{bagel}
C.~Deng, D.~Zhu, K.~Li, C.~Gou, F.~Li, Z.~Wang, S.~Zhong, W.~Yu, X.~Nie, Z.~Song \emph{et~al.}, ``Emerging properties in unified multimodal pretraining,'' \emph{arXiv preprint arXiv:2505.14683}, 2025.

\bibitem{wu2025janus}
C.~Wu, X.~Chen, Z.~Wu, Y.~Ma, X.~Liu, Z.~Pan, W.~Liu, Z.~Xie, X.~Yu, C.~Ruan \emph{et~al.}, ``Janus: Decoupling visual encoding for unified multimodal understanding and generation,'' in \emph{Proceedings of the Computer Vision and Pattern Recognition Conference}, 2025, pp. 12\,966--12\,977.

\bibitem{xie2025show}
J.~Xie, Z.~Yang, and M.~Z. Shou, ``Show-o2: Improved native unified multimodal models,'' \emph{arXiv preprint arXiv:2506.15564}, 2025.

\bibitem{unitok}
C.~Ma, Y.~Jiang, J.~Wu, J.~Yang, X.~Yu, Z.~Yuan, B.~Peng, and X.~Qi, ``Unitok: A unified tokenizer for visual generation and understanding,'' \emph{arXiv preprint arXiv:2502.20321}, 2025.

\bibitem{wu2025harmonizing}
S.~Wu, W.~Zhang, L.~Xu, S.~Jin, Z.~Wu, Q.~Tao, W.~Liu, W.~Li, and C.~C. Loy, ``Harmonizing visual representations for unified multimodal understanding and generation,'' \emph{arXiv preprint arXiv:2503.21979}, 2025.

\bibitem{cao2025hunyuanimage}
S.~Cao, H.~Chen, P.~Chen, Y.~Cheng, Y.~Cui, X.~Deng, Y.~Dong, K.~Gong, T.~Gu, X.~Gu \emph{et~al.}, ``Hunyuanimage 3.0 technical report,'' \emph{arXiv preprint arXiv:2509.23951}, 2025.

\bibitem{mogao}
C.~Liao, L.~Liu, X.~Wang, Z.~Luo, X.~Zhang, W.~Zhao, J.~Wu, L.~Li, Z.~Tian, and W.~Huang, ``Mogao: An omni foundation model for interleaved multi-modal generation,'' \emph{arXiv preprint arXiv:2505.05472}, 2025.

\bibitem{zhou2024transfusion}
C.~Zhou, L.~Yu, A.~Babu, K.~Tirumala, M.~Yasunaga, L.~Shamis, J.~Kahn, X.~Ma, L.~Zettlemoyer, and O.~Levy, ``Transfusion: Predict the next token and diffuse images with one multi-modal model,'' \emph{arXiv preprint arXiv:2408.11039}, 2024.

\bibitem{seed-x}
Y.~Ge, S.~Zhao, J.~Zhu, Y.~Ge, K.~Yi, L.~Song, C.~Li, X.~Ding, and Y.~Shan, ``Seed-x: Multimodal models with unified multi-granularity comprehension and generation,'' \emph{arXiv preprint arXiv:2404.14396}, 2024.

\bibitem{unilip}
H.~Tang, C.~Xie, X.~Bao, T.~Weng, P.~Li, Y.~Zheng, and L.~Wang, ``Unilip: Adapting clip for unified multimodal understanding, generation and editing,'' \emph{arXiv preprint arXiv:2507.23278}, 2025.

\bibitem{blip3-o}
J.~Chen, Z.~Xu, X.~Pan, Y.~Hu, C.~Qin, T.~Goldstein, L.~Huang, T.~Zhou, S.~Xie, S.~Savarese \emph{et~al.}, ``Blip3-o: A family of fully open unified multimodal models-architecture, training and dataset,'' \emph{arXiv preprint arXiv:2505.09568}, 2025.

\bibitem{clip}
A.~Radford, J.~W. Kim, C.~Hallacy, A.~Ramesh, G.~Goh, S.~Agarwal, G.~Sastry, A.~Askell, P.~Mishkin, J.~Clark \emph{et~al.}, ``Learning transferable visual models from natural language supervision,'' in \emph{International conference on machine learning}.\hskip 1em plus 0.5em minus 0.4em\relax PmLR, 2021, pp. 8748--8763.

\bibitem{vae}
D.~P. Kingma, M.~Welling \emph{et~al.}, ``An introduction to variational autoencoders,'' \emph{Foundations and Trends{\textregistered} in Machine Learning}, vol.~12, no.~4, pp. 307--392, 2019.

\bibitem{liu2025tuna}
Z.~Liu, W.~Ren, H.~Liu, Z.~Zhou, S.~Chen, H.~Qiu, X.~Huang, Z.~An, F.~Yang, A.~Patel \emph{et~al.}, ``Tuna: Taming unified visual representations for native unified multimodal models,'' \emph{arXiv preprint arXiv:2512.02014}, 2025.

\bibitem{jiao2025unitoken}
Y.~Jiao, H.~Qiu, Z.~Jie, S.~Chen, J.~Chen, L.~Ma, and Y.-G. Jiang, ``Unitoken: Harmonizing multimodal understanding and generation through unified visual encoding,'' in \emph{Proceedings of the Computer Vision and Pattern Recognition Conference}, 2025, pp. 3600--3610.

\bibitem{internvl2.5}
Z.~Chen, W.~Wang, Y.~Cao, Y.~Liu, Z.~Gao, E.~Cui, J.~Zhu, S.~Ye, H.~Tian, Z.~Liu \emph{et~al.}, ``Expanding performance boundaries of open-source multimodal models with model, data, and test-time scaling,'' \emph{arXiv preprint arXiv:2412.05271}, 2024.

\bibitem{internvl}
Z.~Chen, J.~Wu, W.~Wang, W.~Su, G.~Chen, S.~Xing, M.~Zhong, Q.~Zhang, X.~Zhu, L.~Lu \emph{et~al.}, ``Internvl: Scaling up vision foundation models and aligning for generic visual-linguistic tasks,'' in \emph{Proceedings of the IEEE/CVF Conference on Computer Vision and Pattern Recognition}, 2024, pp. 24\,185--24\,198.

\bibitem{ghosh2023geneval}
D.~Ghosh, H.~Hajishirzi, and L.~Schmidt, ``Geneval: An object-focused framework for evaluating text-to-image alignment,'' \emph{Advances in Neural Information Processing Systems}, vol.~36, pp. 52\,132--52\,152, 2023.

\bibitem{hu2024ella}
X.~Hu, R.~Wang, Y.~Fang, B.~Fu, P.~Cheng, and G.~Yu, ``Ella: Equip diffusion models with llm for enhanced semantic alignment,'' \emph{arXiv preprint arXiv:2403.05135}, 2024.

\bibitem{niu2025wise}
Y.~Niu, M.~Ning, M.~Zheng, W.~Jin, B.~Lin, P.~Jin, J.~Liao, C.~Feng, K.~Ning, B.~Zhu \emph{et~al.}, ``Wise: A world knowledge-informed semantic evaluation for text-to-image generation,'' \emph{arXiv preprint arXiv:2503.07265}, 2025.

\bibitem{dit}
W.~Peebles and S.~Xie, ``Scalable diffusion models with transformers,'' in \emph{Proceedings of the IEEE/CVF international conference on computer vision}, 2023, pp. 4195--4205.

\bibitem{bai2023qwen}
J.~Bai, S.~Bai, Y.~Chu, Z.~Cui, K.~Dang, X.~Deng, Y.~Fan, W.~Ge, Y.~Han, F.~Huang \emph{et~al.}, ``Qwen technical report,'' \emph{arXiv preprint arXiv:2309.16609}, 2023.

\bibitem{liu2023visual}
H.~Liu, C.~Li, Q.~Wu, and Y.~J. Lee, ``Visual instruction tuning,'' \emph{Advances in neural information processing systems}, vol.~36, pp. 34\,892--34\,916, 2023.

\bibitem{li2024llava}
B.~Li, Y.~Zhang, D.~Guo, R.~Zhang, F.~Li, H.~Zhang, K.~Zhang, P.~Zhang, Y.~Li, Z.~Liu \emph{et~al.}, ``Llava-onevision: Easy visual task transfer,'' \emph{arXiv preprint arXiv:2408.03326}, 2024.

\bibitem{qu2025tokenflow}
L.~Qu, H.~Zhang, Y.~Liu, X.~Wang, Y.~Jiang, Y.~Gao, H.~Ye, D.~K. Du, Z.~Yuan, and X.~Wu, ``Tokenflow: Unified image tokenizer for multimodal understanding and generation,'' in \emph{Proceedings of the Computer Vision and Pattern Recognition Conference}, 2025, pp. 2545--2555.

\bibitem{xie2024show}
J.~Xie, W.~Mao, Z.~Bai, D.~J. Zhang, W.~Wang, K.~Q. Lin, Y.~Gu, Z.~Chen, Z.~Yang, and M.~Z. Shou, ``Show-o: One single transformer to unify multimodal understanding and generation,'' \emph{arXiv preprint arXiv:2408.12528}, 2024.

\bibitem{ma2025janusflow}
Y.~Ma, X.~Liu, X.~Chen, W.~Liu, C.~Wu, Z.~Wu, Z.~Pan, Z.~Xie, H.~Zhang, X.~Yu \emph{et~al.}, ``Janusflow: Harmonizing autoregression and rectified flow for unified multimodal understanding and generation,'' in \emph{Proceedings of the Computer Vision and Pattern Recognition Conference}, 2025, pp. 7739--7751.

\bibitem{chen2025janus}
X.~Chen, Z.~Wu, X.~Liu, Z.~Pan, W.~Liu, Z.~Xie, X.~Yu, and C.~Ruan, ``Janus-pro: Unified multimodal understanding and generation with data and model scaling,'' \emph{arXiv preprint arXiv:2501.17811}, 2025.

\bibitem{huang2025ming}
Z.~Huang, D.~Zheng, C.~Zou, R.~Liu, X.~Wang, K.~Ji, W.~Chai, J.~Sun, L.~Wang, Y.~Lv \emph{et~al.}, ``Ming-univision: Joint image understanding and generation with a unified continuous tokenizer,'' \emph{arXiv preprint arXiv:2510.06590}, 2025.

\bibitem{muse-vl}
R.~Xie, C.~Du, P.~Song, and C.~Liu, ``Muse-vl: Modeling unified vlm through semantic discrete encoding,'' in \emph{Proceedings of the IEEE/CVF International Conference on Computer Vision}, 2025, pp. 24\,135--24\,146.

\bibitem{wu2024vila}
Y.~Wu, Z.~Zhang, J.~Chen, H.~Tang, D.~Li, Y.~Fang, L.~Zhu, E.~Xie, H.~Yin, L.~Yi \emph{et~al.}, ``Vila-u: a unified foundation model integrating visual understanding and generation,'' \emph{arXiv preprint arXiv:2409.04429}, 2024.

\bibitem{qwen2.5}
A.~Yang, B.~Yang, B.~Zhang, B.~Hui, B.~Zheng, B.~Yu, C.~Li, D.~Liu, F.~Huang, H.~Wei, H.~Lin, J.~Yang, J.~Tu, J.~Zhang, J.~Yang, J.~Yang, J.~Zhou, J.~Lin, K.~Dang, K.~Lu, K.~Bao, K.~Yang, L.~Yu, M.~Li, M.~Xue, P.~Zhang, Q.~Zhu, R.~Men, R.~Lin, T.~Li, T.~Xia, X.~Ren, X.~Ren, Y.~Fan, Y.~Su, Y.~Zhang, Y.~Wan, Y.~Liu, Z.~Cui, Z.~Zhang, and Z.~Qiu, ``Qwen2.5 technical report,'' \emph{arXiv preprint arXiv:2412.15115}, 2024.

\bibitem{loshchilov2017decoupled}
I.~Loshchilov and F.~Hutter, ``Decoupled weight decay regularization,'' \emph{arXiv preprint arXiv:1711.05101}, 2017.

\bibitem{esser2024scaling}
P.~Esser, S.~Kulal, A.~Blattmann, R.~Entezari, J.~M{\"u}ller, H.~Saini, Y.~Levi, D.~Lorenz, A.~Sauer, F.~Boesel \emph{et~al.}, ``Scaling rectified flow transformers for high-resolution image synthesis,'' in \emph{Forty-first international conference on machine learning}, 2024.

\bibitem{flux}
B.~F. Labs, S.~Batifol, A.~Blattmann, F.~Boesel, S.~Consul, C.~Diagne, T.~Dockhorn, J.~English, Z.~English, P.~Esser \emph{et~al.}, ``Flux. 1 kontext: Flow matching for in-context image generation and editing in latent space,'' \emph{arXiv preprint arXiv:2506.15742}, 2025.

\bibitem{yue2025uniflow}
Z.~Yue, H.~Zhang, X.~Zeng, B.~Chen, C.~Wang, S.~Zhuang, L.~Dong, K.~Du, Y.~Wang, L.~Wang \emph{et~al.}, ``Uniflow: A unified pixel flow tokenizer for visual understanding and generation,'' \emph{arXiv preprint arXiv:2510.10575}, 2025.

\bibitem{sd}
R.~Rombach, A.~Blattmann, D.~Lorenz, P.~Esser, and B.~Ommer, ``High-resolution image synthesis with latent diffusion models,'' in \emph{Proceedings of the IEEE/CVF conference on computer vision and pattern recognition}, 2022, pp. 10\,684--10\,695.

\bibitem{vavae}
J.~Yao, B.~Yang, and X.~Wang, ``Reconstruction vs. generation: Taming optimization dilemma in latent diffusion models,'' in \emph{Proceedings of the Computer Vision and Pattern Recognition Conference}, 2025, pp. 15\,703--15\,712.

\bibitem{flowmo}
A.~Shaulov, I.~Hazan, L.~Wolf, and H.~Chefer, ``Flowmo: Variance-based flow guidance for coherent motion in video generation,'' \emph{arXiv preprint arXiv:2506.01144}, 2025.

\bibitem{qlip}
Y.~Zhao, F.~Xue, S.~Reed, L.~Fan, Y.~Zhu, J.~Kautz, Z.~Yu, P.~Kr{\"a}henb{\"u}hl, and D.-A. Huang, ``Qlip: Text-aligned visual tokenization unifies auto-regressive multimodal understanding and generation,'' \emph{arXiv preprint arXiv:2502.05178}, 2025.

\bibitem{uae}
W.~Fan, H.~Diao, Q.~Wang, D.~Lin, and Z.~Liu, ``The prism hypothesis: Harmonizing semantic and pixel representations via unified autoencoding,'' \emph{arXiv preprint arXiv:2512.19693}, 2025.

\bibitem{rae}
B.~Zheng, N.~Ma, S.~Tong, and S.~Xie, ``Diffusion transformers with representation autoencoders,'' \emph{arXiv preprint arXiv:2510.11690}, 2025.

\bibitem{liu2024ocrbench}
Y.~Liu, Z.~Li, M.~Huang, B.~Yang, W.~Yu, C.~Li, X.-C. Yin, C.-L. Liu, L.~Jin, and X.~Bai, ``Ocrbench: on the hidden mystery of ocr in large multimodal models,'' \emph{Science China Information Sciences}, vol.~67, no.~12, p. 220102, 2024.

\bibitem{chen2024mmstar}
L.~Chen, J.~Li, X.~Dong, P.~Zhang, Y.~Zang, Z.~Chen, H.~Duan, J.~Wang, Y.~Qiao, D.~Lin \emph{et~al.}, ``Are we on the right way for evaluating large vision-language models?'' \emph{Advances in Neural Information Processing Systems}, vol.~37, pp. 27\,056--27\,087, 2024.

\bibitem{li2023seed}
B.~Li, R.~Wang, G.~Wang, Y.~Ge, Y.~Ge, and Y.~Shan, ``Seed-bench: Benchmarking multimodal llms with generative comprehension,'' \emph{arXiv preprint arXiv:2307.16125}, 2023.

\bibitem{zhao2026unicom}
Y.~Zhao, W.~Lin, Z.~Zhang, M.~Yang, J.~Chen, W.~Zhang, Z.~Zhong, and L.~Bo, ``Unicom: Unified multimodal modeling via compressed continuous semantic representations,'' \emph{arXiv preprint arXiv:2603.10702}, 2026.

\bibitem{hu2025omni}
J.~Hu, S.~Zhao, Q.-G. Chen, X.~Qiu, J.~Liu, Z.~Xu, W.~Luo, K.~Zhang, and Y.~Lu, ``Omni-view: Unlocking how generation facilitates understanding in unified 3d model based on multiview images,'' \emph{arXiv preprint arXiv:2511.07222}, 2025.

\bibitem{li2025unif}
J.~Li, S.~Zhou, L.~Guo, X.~Qiu, L.~Xu, D.~Qu, T.~Long, C.~Fan, M.~Li, H.~Fan \emph{et~al.}, ``Uniface: A unified fine-grained face understanding and generation model,'' \emph{arXiv preprint arXiv:2503.08120}, 2025.

\bibitem{tu2024motioneditor}
S.~Tu, Q.~Dai, Z.-Q. Cheng, H.~Hu, X.~Han, Z.~Wu, and Y.-G. Jiang, ``Motioneditor: Editing video motion via content-aware diffusion,'' in \emph{CVPR}, 2024.

\bibitem{tu2026baton}
S.~Tu, Q.~Tian, Z.~Yang, Y.~Wu, X.~Han, W.~Kong, J.~Xiong, J.-W. Zhang, Z.~Zhong, L.~Bo \emph{et~al.}, ``Baton: Explicit semantic blueprints for joint video-audio generation,'' \emph{arXiv preprint arXiv:2605.25195}, 2026.

\bibitem{huang2026agentic}
Y.~Huang, S.~Tu, X.~Yan, J.~Zhan, Z.~Yang, Z.~Xing, H.~Zhang, T.~Zhang, Y.-G. Jiang, and Z.~Wu, ``Agentic visual generation: From generative models to agentic control,'' \emph{arXiv preprint arXiv:2609.06758}, 2026.

\bibitem{huang2026va}
Y.~Huang, S.~Tu, X.~Yan, Z.~Yang, J.~Han, X.~Hang, Y.-G. Jiang, and Z.~Wu, ``Va-judger: Reward modeling from human preference feedback for joint video-audio generation,'' \emph{arXiv preprint arXiv:2608.18607}, 2026.

\bibitem{pan2025transfer}
X.~Pan, S.~N. Shukla, A.~Singh, Z.~Zhao, S.~K. Mishra, J.~Wang, Z.~Xu, J.~Chen, K.~Li, F.~Juefei-Xu \emph{et~al.}, ``Transfer between modalities with metaqueries,'' \emph{arXiv preprint arXiv:2504.06256}, 2025.

\bibitem{wu2025openuni}
S.~Wu, Z.~Wu, Z.~Gong, Q.~Tao, S.~Jin, Q.~Li, W.~Li, and C.~C. Loy, ``Openuni: A simple baseline for unified multimodal understanding and generation,'' \emph{arXiv preprint arXiv:2505.23661}, 2025.

\bibitem{shi2025latent}
M.~Shi, H.~Wang, W.~Zheng, Z.~Yuan, X.~Wu, X.~Wang, P.~Wan, J.~Zhou, and J.~Lu, ``Latent diffusion model without variational autoencoder,'' \emph{arXiv preprint arXiv:2510.15301}, 2025.

\bibitem{yao2025vtp}
J.~Yao, Y.~Song, Y.~Zhou, and X.~Wang, ``Towards scalable pre-training of visual tokenizers for generation,'' \emph{arXiv preprint arXiv:2512.13687}, 2025.

\bibitem{llava1.5}
H.~Liu, C.~Li, Y.~Li, and Y.~J. Lee, ``Improved baselines with visual instruction tuning,'' 2023.

\bibitem{pope}
Y.~Li, Y.~Du, K.~Zhou, J.~Wang, W.~X. Zhao, and J.-R. Wen, ``Evaluating object hallucination in large vision-language models,'' \emph{arXiv preprint arXiv:2305.10355}, 2023.

\bibitem{mmbench}
Y.~Liu, H.~Duan, Y.~Zhang, B.~Li, S.~Zhang, W.~Zhao, Y.~Yuan, J.~Wang, C.~He, Z.~Liu \emph{et~al.}, ``Mmbench: Is your multi-modal model an all-around player?'' in \emph{European conference on computer vision}.\hskip 1em plus 0.5em minus 0.4em\relax Springer, 2024, pp. 216--233.

\bibitem{yue2024mmmu}
X.~Yue, Y.~Ni, K.~Zhang, T.~Zheng, R.~Liu, G.~Zhang, S.~Stevens, D.~Jiang, W.~Ren, Y.~Sun \emph{et~al.}, ``Mmmu: A massive multi-discipline multimodal understanding and reasoning benchmark for expert agi,'' in \emph{Proceedings of the IEEE/CVF Conference on Computer Vision and Pattern Recognition}, 2024, pp. 9556--9567.

\bibitem{kembhavi2016ai2d}
A.~Kembhavi, M.~Salvato, E.~Kolve, M.~Seo, H.~Hajishirzi, and A.~Farhadi, ``A diagram is worth a dozen images,'' in \emph{European conference on computer vision}.\hskip 1em plus 0.5em minus 0.4em\relax Springer, 2016, pp. 235--251.

\bibitem{nitro}
T.~Shen, J.~Yu, D.~Zhou, D.~Li, and E.~Barsoum, ``E-mmdit: Revisiting multimodal diffusion transformer design for fast image synthesis under limited resources,'' \emph{arXiv preprint arXiv:2510.27135}, 2025.

\bibitem{sam}
A.~Kirillov, E.~Mintun, N.~Ravi, H.~Mao, C.~Rolland, L.~Gustafson, T.~Xiao, S.~Whitehead, A.~C. Berg, W.-Y. Lo \emph{et~al.}, ``Segment anything,'' in \emph{Proceedings of the IEEE/CVF international conference on computer vision}, 2023, pp. 4015--4026.

\bibitem{sun2023journeydb}
K.~Sun, J.~Pan, Y.~Ge, H.~Li, H.~Duan, X.~Wu, R.~Zhang, A.~Zhou, Z.~Qin, Y.~Wang \emph{et~al.}, ``Journeydb: A benchmark for generative image understanding,'' \emph{Advances in neural information processing systems}, vol.~36, pp. 49\,659--49\,678, 2023.

\bibitem{imagenet}
O.~Russakovsky, J.~Deng, H.~Su, J.~Krause, S.~Satheesh, S.~Ma, Z.~Huang, A.~Karpathy, A.~Khosla, M.~S. Bernstein, A.~C. Berg, and L.~Fei-Fei, ``Imagenet large scale visual recognition challenge,'' \emph{International Journal of Computer Vision}, vol. 115, pp. 211 -- 252, 2014.

\bibitem{gadre2023datacomp}
S.~Y. Gadre, G.~Ilharco, A.~Fang, J.~Hayase, G.~Smyrnis, T.~Nguyen, R.~Marten, M.~Wortsman, D.~Ghosh, J.~Zhang \emph{et~al.}, ``Datacomp: In search of the next generation of multimodal datasets,'' \emph{Advances in Neural Information Processing Systems}, vol.~36, pp. 27\,092--27\,112, 2023.

\bibitem{kakaobrain2022coyo-700m}
M.~Byeon, B.~Park, H.~Kim, S.~Lee, W.~Baek, and S.~Kim, ``Coyo-700m: Image-text pair dataset,'' \url{https://github.com/kakaobrain/coyo-dataset}, 2022.

\bibitem{li2024densefusion}
X.~Li, F.~Zhang, H.~Diao, Y.~Wang, X.~Wang, and L.~Duan, ``Densefusion-1m: Merging vision experts for comprehensive multimodal perception,'' \emph{Advances in Neural Information Processing Systems}, vol.~37, pp. 18\,535--18\,556, 2024.

\bibitem{Llava-onevision}
B.~Li, Y.~Zhang, D.~Guo, R.~Zhang, F.~Li, H.~Zhang, K.~Zhang, P.~Zhang, Y.~Li, Z.~Liu \emph{et~al.}, ``Llava-onevision: Easy visual task transfer,'' \emph{arXiv preprint arXiv:2408.03326}, 2024.

\bibitem{changpinyo2021conceptual}
S.~Changpinyo, P.~Sharma, N.~Ding, and R.~Soricut, ``Conceptual 12m: Pushing web-scale image-text pre-training to recognize long-tail visual concepts,'' in \emph{Proceedings of the IEEE/CVF conference on computer vision and pattern recognition}, 2021, pp. 3558--3568.

\bibitem{dalle}
A.~Ramesh, M.~Pavlov, G.~Goh, S.~Gray, C.~Voss, A.~Radford, M.~Chen, and I.~Sutskever, ``Zero-shot text-to-image generation,'' \emph{ArXiv}, vol. abs/2102.12092, 2021.

\bibitem{wang2024emu3}
X.~Wang, X.~Zhang, Z.~Luo, Q.~Sun, Y.~Cui, J.~Wang, F.~Zhang, Y.~Wang, Z.~Li, Q.~Yu \emph{et~al.}, ``Emu3: Next-token prediction is all you need,'' \emph{arXiv preprint arXiv:2409.18869}, 2024.

\bibitem{mme}
C.~Fu, P.~Chen, Y.~Shen, Y.~Qin, M.~Zhang, X.~Lin, J.~Yang, X.~Zheng, K.~Li, X.~Sun \emph{et~al.}, ``Mme: A comprehensive evaluation benchmark for multimodal large language models,'' \emph{arXiv preprint arXiv:2306.13394}, 2023.

\bibitem{sdxl}
D.~Podell, Z.~English, K.~Lacey, A.~Blattmann, T.~Dockhorn, J.~M{\"u}ller, J.~Penna, and R.~Rombach, ``Sdxl: Improving latent diffusion models for high-resolution image synthesis,'' \emph{arXiv preprint arXiv:2307.01952}, 2023.

\end{thebibliography}

\newpage
\appendix

\section{Related Works}
\label{sec:related_works}
\subsection{Unified Multimodal Models}
The pursuit of a singular architecture for both visual understanding and generation has led to the evolution of Unified Multimodal Models (UMMs) \cite{bagel,zhao2026unicom,hu2025omni,li2025unif}, broadly categorized into composite and native architectures. Complementary generation-focused studies explore various application ~\cite{tu2024motioneditor,tu2026baton,huang2026agentic,huang2026va}.

Composite UMMs such as MetaQuery~\cite{pan2025transfer}, BLIP-3o~\cite{blip3-o}, OpenUni \cite{wu2025openuni} and Unilip~\cite{unilip} achieve capability extension by integrating specialized, pre-trained experts (e.g., separate LLMs and Diffusion models) via lightweight adapters. While effective for rapid deployment, their reliance on frozen backbones restricts deep inter-modal interaction, resulting in a shallow unification where understanding and generation remain functionally decoupled.

Native UMMs strive for deeper integration by training holistically within a shared parameter space. However, as analyzed in Fig. \ref{fig:intro}, existing methods are constrained by a fundamental representation trilemma: (i) {Decoupled Architectures}, such as the Janus series~\cite{wu2025janus,ma2025janusflow}, Bagel \cite{bagel} and UniToken~\cite{jiao2025unitoken}, mitigate task interference by utilizing separate or dual encoders for vision and language. While UniToken concatenates discrete and continuous tokens to bridge modalities, this dual-encoder design inherently sacrifices \textit{Unification of Input Representation}, leading to parameter redundancy, computational inefficiency due to sequence inflation, and severed task synergy.
(ii) {Sequential Architectures}, including Show-o2~\cite{xie2025show} and TUNA~\cite{liu2025tuna}, attempt to unify inputs by stacking semantic encoders atop a VAE. However, the aggressive spatial downsampling required by the VAE acts as a lossy information bottleneck, disrupting the \textit{Coherence of Information Flow} and irreversibly diluting the fine-grained structural signals needed for generation.
(iii) {Shared Architectures}, such as Unitok~\cite{unitok}, Transfusion~\cite{zhou2024transfusion}, and Harmon~\cite{wu2025harmonizing}, attempt to unify tasks but struggle with the \textit{Compatibility of Learning Process}. 
On one hand, discrete approaches like Unitok \cite{unitok} suffer from quantization errors inherent in codebooks, which fundamentally conflict with the high-fidelity requirements of synthesis. 
On the other hand, continuous approaches like Transfusion \cite{zhou2024transfusion} and Harmon \cite{wu2025harmonizing} employ VAE latent features for both tasks; while these representations are adequate for generation, they lack the {high-dimensional semantic abstractions} required for perception, resulting in suboptimal performance on complex understanding tasks.
In contrast, our \model achieves rational unification by resolving these conflicts through a harmonized, continuous representation.

\subsection{Unified Tokenizers and Representations}
Recent research has shifted from discrete to continuous unified representations to bridge the gap between semantic abstraction and pixel-level fidelity. To circumvent traditional convolutional or pre-trained VAEs, paradigms such as RAE~\cite{rae} and SVG~\cite{shi2025latent} employ frozen, pre-trained semantic encoders. While this accelerates convergence, the inherent loss of low-level features compromises reconstruction fidelity. Although UniFlow~\cite{yue2025uniflow} attempts to recover structural details via self-distillation, it relies on dual-stream features, failing to achieve a truly unified, single-stream information flow. Similarly, QLIP~\cite{qlip} seeks to unify representations by aligning visual tokens with textual semantics; however, this heavy reliance on linguistic alignment imposes a semantic bottleneck, filtering out the high-frequency textural details necessary for photorealistic synthesis. Furthermore, while VTP~\cite{yao2025vtp} improves generative scaling by jointly optimizing contrastive and reconstruction losses, its low-dimensional latent space is tailored for generation and lacks the representational capacity required for complex multimodal perception. Distinctly, our \tokenizer employs a {functionally progressive} pure ViT backbone  to actively bridge the representation gap. This design ensures a coherent single-stream flow, eliminating quantization errors while simultaneously satisfying the conflicting demands of high-fidelity synthesis and deep semantic understanding.

\section{Limitations and Future Work}
Despite establishing promising benchmarks, our current implementation faces limitations in scale that define future research directions. A primary constraint is the 300M parameter \tokenizer, which creates a potential bottleneck for encoding intricate details in fine-grained tasks. Furthermore, relying on a 7B parameter model trained on only 100M image-text pairs restricts generalizability and world knowledge absorption compared to state-of-the-art foundation models. Consequently, our immediate future work will center on significantly scaling both the model architecture and the training dataset size to address these gaps and unlock robust capabilities across diverse multimodal tasks.

\section{Broader Impacts}
As \model introduces state-of-the-art text-to-image generation capabilities within a unified framework, we acknowledge potential downstream risks. These include the generation of misleading or fabricated visual content (e.g., deepfakes), which could be exploited for disinformation or impersonation. To mitigate such risks, we advocate for the incorporation of content watermarking and provenance tracking mechanisms upon deployment, as well as adherence to responsible release practices such as gated model access and usage guidelines. We believe that advancing the scientific understanding of unified multimodal architectures carries substantial positive societal value, while the associated risks can be effectively managed through community-driven safety standards.

\section{Training  details}
\label{training:details}
\paragraph{\tokenizer}  is training in two progressive stages to effectively balance representation learning and generative quality. Stage 1: Foundation Training. In the initial stage, we focus on establishing the foundational capabilities of the tokenizer through joint reconstruction and distillation objectives. We utilize a large-scale composite dataset consisting of ImageNet-1.2M, CC-12M, and SAM-10M. During this phase, the entire tokenizer (both encoder and decoder) is trained end-to-end. We optimize the model for a total of 300k iterations with a global batch size of 256. The learning rate is set to $2e-5$. Stage 2: Decoder Refinement. In the second stage, we shift our focus to refining the fidelity of the generated images. We freeze the parameters of the pre-trained encoder to preserve the learned semantic representations and exclusively fine-tune the decoder. In this phase, the distillation objective is discarded. Instead, we incorporate adversarial training (GAN loss) alongside the reconstruction loss to enhance perceptual quality. The learning rate is reduced to $1e\text{-}5$, and the training proceeds for an additional 100k iterations.

\par
 \paragraph{\model} training proceeds through a progressive three-stage curriculum, with specific hyperparameters detailed in Tab. ~\ref{tab:training_details}. Across all three stages, we maintain a constant input image resolution of $448 \times 448$ and a global batch size of 1024. We employ varying learning rates (LRs) tailored to different model components and training phases. In Stage I, the Vision Head and Sem-ViT are trained with LRs of $10^{-4}$ and $5\times10^{-5}$, respectively. Progressing to Stages II and III, the Vision Head LR is reduced to $5\times10^{-5}$, while the LLM and Sem-ViT are trained jointly at a lower rate of $2\times10^{-5}$. The training strategy also involves shifting data mixing ratios (Text Understanding : Image Understanding : Image Generation) and durations: Stage I utilizes a data ratio of 0:1:3 for 150K steps; Stage II adjusts the ratio to 0:2:2 for 100K steps; and Stage III concludes with a ratio of 1:2:2 for 20K steps.
\begin{table}[htbp]
    \centering
    \caption{\textbf{Training details and computational cost of our \model.} The total training of the UMM stages takes $\sim$72h on 128 H20 GPUs. The \tokenizer pre-training takes an additional 24h on 64 H20 GPUs (totaling $\sim$96h). \textsuperscript{$\dagger$}Data Ratio denotes Text Understanding : Image Understanding : Image Generation.}
    \begin{adjustbox}{max width=0.95\linewidth} 
\begin{tabular}{cccc}
\toprule
Setting & Stage I & Stage II &  Stage III \\
\midrule
 & Vision Head: $10^{-4}$ & Vision Head: $ 5\times10^{-5}$ & Vision Head: $ 5\times10^{-5}$ \\
\multirow{-2}[0]{*}{LR.} & Sem-ViT: $ 5\times10^{-5}$  & LLM \& Sem-ViT: $ 2\times10^{-5}$ & LLM \& Sem-ViT: $ 2\times10^{-5}$ \\
Image Size & 448 & 448 & 448 \\
Batch Size & 1024 & 1024 & 1024 \\
Data Ratio\textsuperscript{$\dagger$} & 0:1:3 & 0:2:2 & 1:2:2  \\
\midrule
Hardware & 128 H20 GPUs & 128 H20 GPUs & 128 H20 GPUs \\
Training Step & 150K & 100K & 20K \\
Time Cost & $\sim$30h & $\sim$30h & $\sim$12h \\
\bottomrule
\end{tabular}
\end{adjustbox}
    \label{tab:training_details}
\end{table}
\subsection{Ablation study training details }
\label{subsec:training_data_abla}
In our ablation studies, the evaluation covers three core capabilities with specific setups: 
(i) {Multimodal Understanding:} Following the LLaVA-1.5 training protocol~\cite{llava1.5}, we report the average score (Avg QA) across POPE~\cite{pope}, MMBench~\cite{mmbench}, MMMU~\cite{yue2024mmmu}, AI2D~\cite{kembhavi2016ai2d}, and RealWorldQA. 
(ii) {Image Generation:} We adapt the Nitro-E framework \cite{nitro} by replacing its VAE with our \tokenizer. We train this setup for 15k iterations using the SAM-1B \cite{sam} and JourneyDB \cite{sun2023journeydb} datasets and evaluate performance on GenEval. 
(iii) {Image Reconstruction:} We train on the ImageNet-1k (1.2M) dataset \cite{imagenet} for 150k iterations and assess quality using rFID.

\subsection{Representation-harmonized co-promotion experiment training details  }
\label{harmon_details}
\begin{figure*}[!t]
    \centering
    \begin{minipage}{0.32\textwidth}
        \centering
        \includegraphics[width=\linewidth]{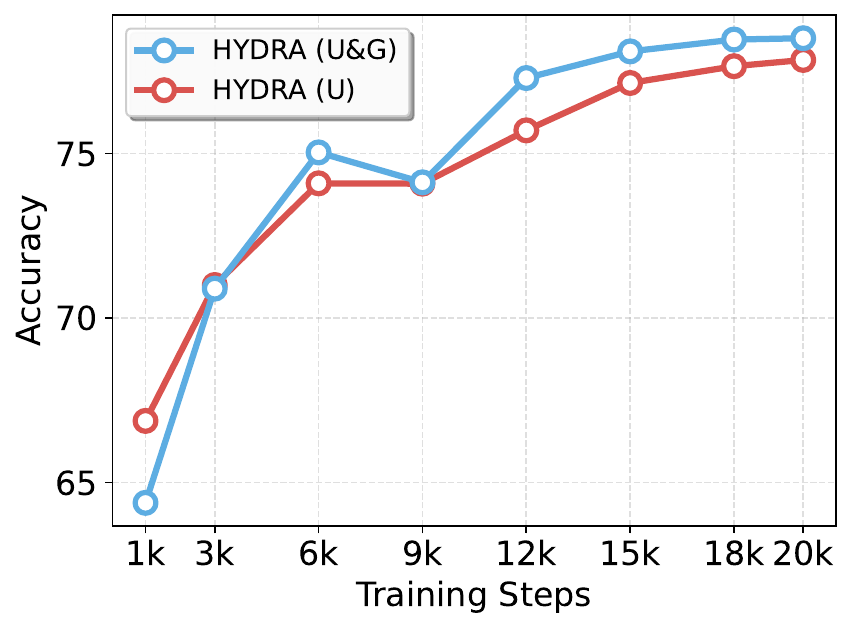}
        \subcaption{AI2D \cite{kembhavi2016ai2d}.} 
        \label{fig:ai2d}
    \end{minipage}\hfill
    \begin{minipage}{0.32\textwidth}
        \centering
        \includegraphics[width=\linewidth]{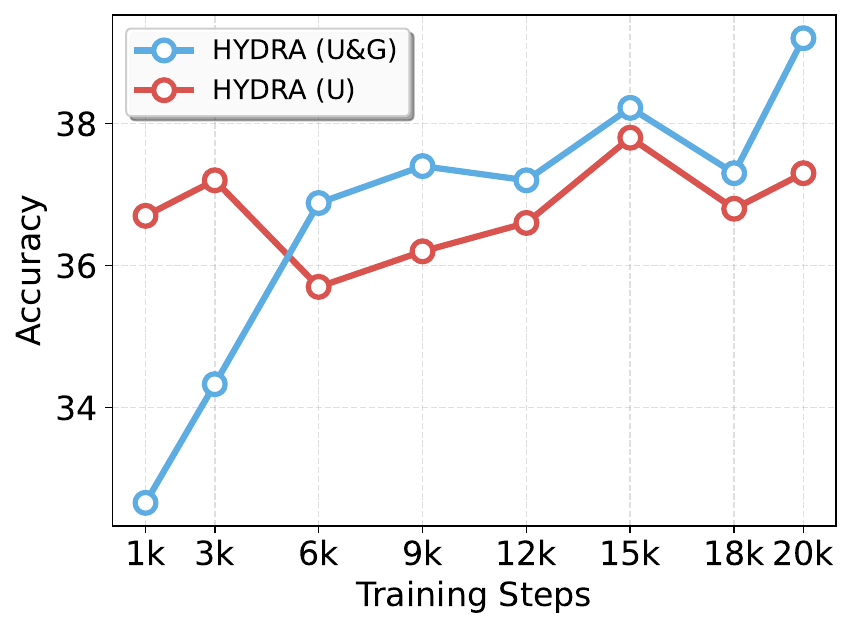}
        \subcaption{MMMU \cite{yue2024mmmu}.} 
        \label{fig:mmu}
    \end{minipage}\hfill
    \begin{minipage}{0.32\textwidth}
        \centering
        \includegraphics[width=\linewidth]{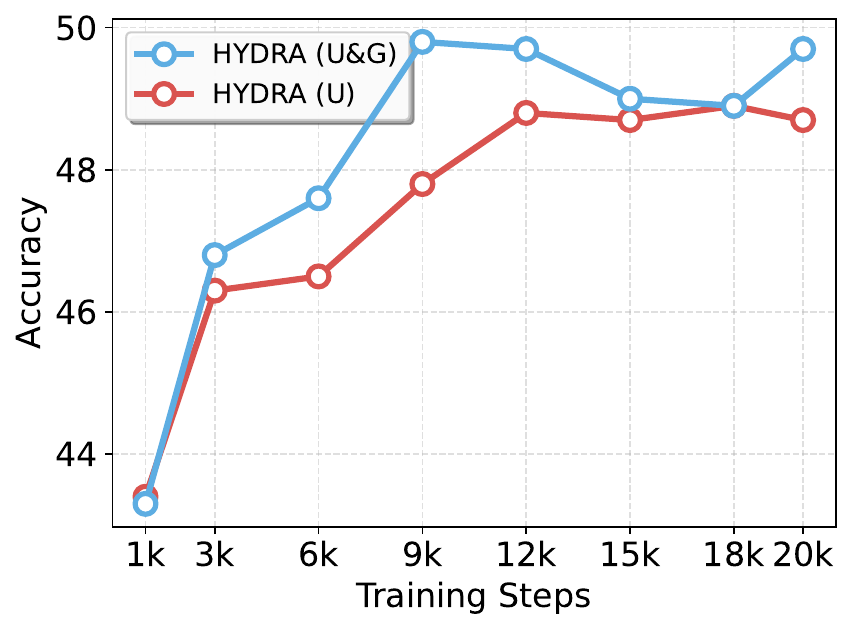}
        \subcaption{OCRBench \cite{liu2024ocrbench}.} 
        \label{fig:ocrbench}
    \end{minipage}
    
    \vspace{1em} 

    \begin{minipage}{0.32\textwidth}
        \centering
        \includegraphics[width=\linewidth]{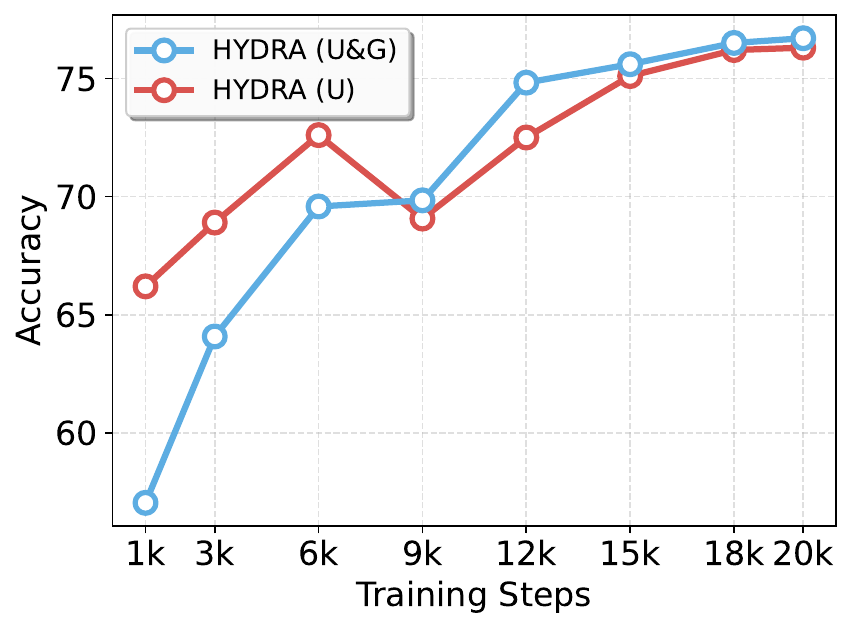}
        \subcaption{MMBench \cite{mmbench}.} 
        \label{fig:mmb}
    \end{minipage}\hfill
    \begin{minipage}{0.32\textwidth}
        \centering
        \includegraphics[width=\linewidth]{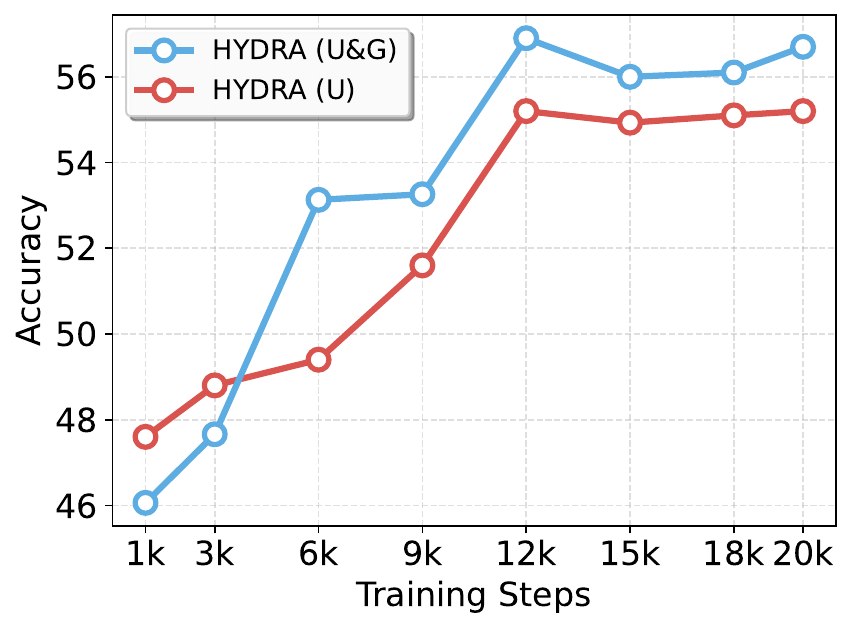}
        \subcaption{MMStar \cite{chen2024mmstar}.} 
        \label{fig:mmstar}
    \end{minipage}\hfill
    \begin{minipage}{0.32\textwidth}
        \centering
        \includegraphics[width=\linewidth]{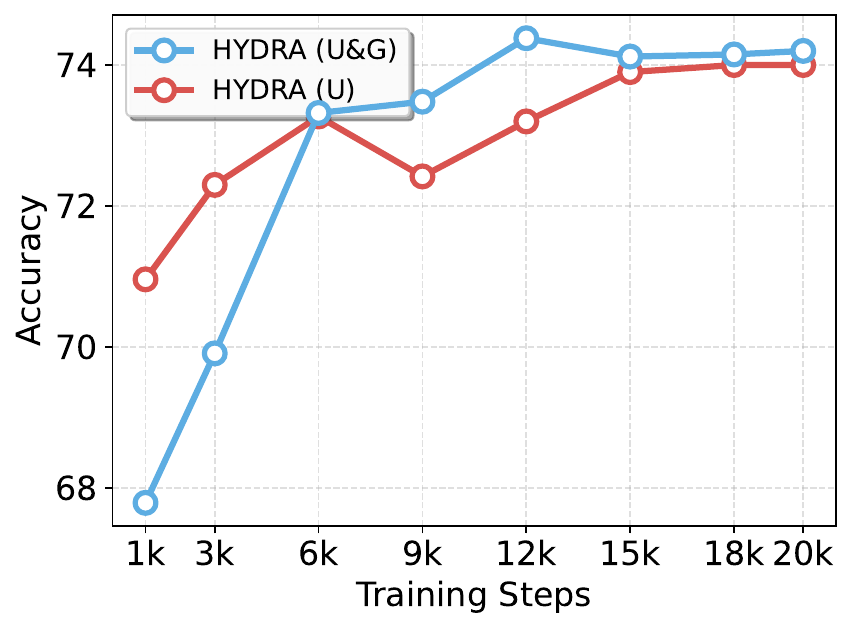}
        \subcaption{SEED-Bench \cite{li2023seed}.} 
        \label{fig:seed}
    \end{minipage}

    \caption{Performance comparison on various benchmarks across different training steps. The benchmarks include general perception (SEED-Bench), reasoning (MMStar, MMMU, MMBench, AI2D), and OCR-related tasks (OCRBench).}
    \label{fig:benchmark_comparison}
\end{figure*}

To validate the mutual benefits of our harmonized representation, we conducted comparative experiments on both generation and understanding tasks. For generation, we trained both Show-o2 and our \model on the same 100M dataset for 150K steps, comparing the GenEval performance of joint training against a generation-only baseline across different training stages. For understanding, we performed Stage III training for 20K steps, comparing joint training with an understanding-only approach. We evaluated the models on general understanding (SEED-Bench), reasoning (MMStar, MMMU, MMBench, AI2D), and OCR-related tasks (OCRBench). Detailed settings for each benchmark are illustrated in Fig. \ref{fig:benchmark_comparison}.

\subsection{Training data}
\label{subsec:training_data}
\subsubsection{Image Understanding}

\paragraph{Stage I.} For the initial pre-training stage, we utilized a massive dataset comprising approximately 22M images paired with captions. These data were sourced from a diverse collection of large-scale datasets, including ImageNet \cite{imagenet}, DataComp-1B \cite{gadre2023datacomp}, COYO-700M \cite{kakaobrain2022coyo-700m}, SA-1B \cite{sam}, and DenseFusion \cite{li2024densefusion}.

\paragraph{Stage II.} In Stage II, we continued to utilize the comprehensive collection of data from the sources employed in Stage I (ImageNet \cite{imagenet}, DataComp-1B \cite{gadre2023datacomp}, COYO-700M \cite{kakaobrain2022coyo-700m}, SA-1B \cite{sam}, and DenseFusion \cite{li2024densefusion}) for further training.

\paragraph{Stage III.} For the final high-quality fine-tuning stage, we directly utilized the refined 3.2M instruction-tuning data from LLaVA-One-Vision~\cite{Llava-onevision} to maximize image understanding performance.

\subsubsection{Image Generation}

\paragraph{Stage I.} For the initial generation training stage, we curated a large-scale dataset totaling approximately {78M} samples. This comprises SA-1B \cite{sam}, CC-12M \cite{changpinyo2021conceptual}, 1M images from DenseFusion \cite{li2024densefusion}, and a caption dataset consisting of {55M} real images.

\paragraph{Stage II.} Stage II expands upon the data used in Stage I. In addition to SA-1B \cite{sam}, CC-12M \cite{changpinyo2021conceptual}, the {55M}  real images, and DenseFusion-1M \cite{li2024densefusion}, we incorporated 10M synthetic images generated through Flux distillation, following the methodology described in~\cite{nitro}.

\paragraph{Stage III.} For high-quality generation fine-tuning, we utilized the JourneyDB dataset \cite{sun2023journeydb}  and DenseFusion-1M \cite{li2024densefusion}. Additionally, we refined the 10M Flux-distilled data from Stage II through filtering, selecting a high-quality subset of 5M images for this final stage.

\section{Additional main results}

\begin{table*}[t]
\centering
\caption{\textbf{Detailed image generation results on the \textbf{GenEval} benchmark.} $^\dagger$ refers to fintuning with GPT-4o distilled dataset \cite{blip3-o}.}
\label{tab:geneval_detailed}
\begin{adjustbox}{max width=0.95\linewidth}
\begin{tabular}{@{}lc cccccc c@{}}
\toprule
\textbf{Models} & \textbf{Size} & \textbf{Single Object} & \textbf{Two Objects} & \textbf{Count} & \textbf{Colors} & \textbf{Position} & \textbf{Color Attribute} & \textbf{Overall} \\ \midrule


\multicolumn{9}{c}{\textit{Generation-only Models}} \\
\midrule
SD3-Med \cite{esser2024scaling} & 2B & 0.99 & 0.94 & 0.72 & 0.89 & 0.33 & 0.60 & 0.74 \\
FLUX.1 [Dev] \cite{flux} & 12B & 0.98 & 0.93 & 0.75 & 0.93 & 0.68 & 0.65 & 0.82 \\ 
DALL-E 3 \cite{dalle} & - & 0.96 & 0.87 & 0.47 & 0.83 & 0.43 & 0.45 & 0.67 \\ \midrule


\multicolumn{9}{c}{\textit{Composite Unified Multimodal Models}} \\
\midrule
TokenFlow-XL \cite{qu2025tokenflow} & 14B & 0.95 & 0.60 & 0.41 & 0.81 & 0.16 & 0.24 & 0.55 \\
SEED-X \cite{seed-x} & 17B & 0.97 & 0.58 & 0.26 & 0.80 & 0.19 & 0.14 & 0.49 \\
MetaQuery-XL \cite{pan2025transfer} & 7B & - & - & - & - & - & - & 0.80 \\
Blip3-o $^\dagger$ \cite{blip3-o} & 8B & - & - & - & - & - & - & 0.84 \\ \midrule


\multicolumn{9}{c}{\textit{Smaller Native Unified Multimodal Models}} \\
\midrule
Show-o \cite{xie2024show} & 1.3B & 0.95 & 0.52 & 0.49 & 0.82 & 0.11 & 0.28 & 0.53 \\
Harmon \cite{wu2025harmonizing} & 1.5B & 0.99 & 0.86 & 0.66 & 0.85 & 0.74 & 0.48 & 0.76 \\
Show-o2 \cite{xie2025show} & 1.5B & 0.99 & 0.86 & 0.55 & 0.86 & 0.46 & 0.63 & 0.73 \\
\rowcolor{icmlblue}
\model & 1.5B& \textbf{1.00} & \textbf{0.95} & \textbf{0.75} & \textbf{0.91} & \textbf{0.76} & \textbf{0.77} & \textbf{0.86} \\  
\midrule

\multicolumn{9}{c}{\textit{Larger Native Unified Multimodal Models}} \\
\midrule
Ming-UniVision~\cite{huang2025ming} & 16B & 1.00 & 0.93 & 0.59 & 0.93 & 0.92 & 0.70 & 0.85 \\
BAGEL \cite{bagel} & 14B & 0.98 & 0.95 & 0.84 & 0.95 & 0.78 & 0.77 & 0.88 \\
MUSE-VL \cite{muse-vl}&7B &0.98 & 0.64 & 0.54 & 0.72 & 0.25 & 0.31 & 0.57\\
Janus-Pro \cite{chen2025janus} & 7B & 0.99 & 0.89 & 0.59 & 0.90 & 0.79 & 0.66 & 0.80 \\
Show-o2 \cite{xie2025show} & 7B & 1.00 & 0.87 & 0.58 & 0.92 & 0.52 & 0.62 & 0.76 \\
\rowcolor{icmlblue}
\model & 7B  & \textbf{1.00} & \textbf{0.97} & \textbf{0.68} & \textbf{0.91} & \textbf{0.81} & \textbf{0.80} & \textbf{0.86}\\ 


\bottomrule
\end{tabular}
\end{adjustbox}
\end{table*}
\begin{table*}[h]
\centering
\caption{\textbf{Detailed image generation results on the \textbf{DPG-Bench} benchmark \cite{hu2024ella}.}}
\label{tab:dpg_results_full}
\begin{adjustbox}{max width=0.95\linewidth}
\begin{tabular}{@{}lc cccccc@{}}
\toprule
\textbf{Models} & \textbf{Size} & \textbf{Global} & \textbf{Entity} & \textbf{Attribute} & \textbf{Relation} & \textbf{Other} & \textbf{Overall} \\ \midrule

\multicolumn{8}{c}{\textit{Generation-only Models}} \\
\midrule
FLUX.1 [Dev] \cite{flux} & 12B & 82.10 & 89.50 & 88.70 & 91.10 & 89.40 & 84.00 \\
\midrule
\multicolumn{8}{c}{\textit{Smaller Native Unified Multimodal Models}} \\
\midrule
Show-o2 \cite{xie2025show} & 1.5B & 87.53 & 90.38 & 91.34 & 90.30 & 91.21 & 85.02 \\
\rowcolor{icmlblue}
\model & 1.5B & \textbf{89.65} & \textbf{91.62} & \textbf{90.24} & \textbf{91.59} & \textbf{90.88} & \textbf{85.51} \\ \midrule

\multicolumn{8}{c}{\textit{Larger Native Unified Multimodal Models}} \\
\midrule
Janus-Pro \cite{chen2025janus} & 7B & 86.90 & 88.90 & 89.40 & 89.32 & 89.48 & 84.19 \\
BAGEL \cite{bagel} & 14B & 88.94 & 90.37 & 91.29 & 90.82 & 88.67 & 85.07 \\
Show-o2 \cite{xie2025show} & 7B & 89.00 & 91.78 & 89.96 & 91.81 & 91.64 & 86.14 \\
\rowcolor{icmlblue}
\model & 7B & \textbf{90.22} & \textbf{91.88} & \textbf{89.92} & \textbf{91.83} & \textbf{90.93} & \textbf{86.40} \\
\bottomrule
\end{tabular}
\end{adjustbox}
\end{table*}
\begin{table*}[t]
\centering
\caption{\textbf{Detailed image generation results on the \textbf{WISE} benchmark \cite{niu2025wise}.}}
\label{tab:wise_detailed}
\begin{adjustbox}{max width=0.95\linewidth}
\begin{tabular}{@{}lc ccccccc@{}}
\toprule
\textbf{Models} & \textbf{Size} & \textbf{Culture} & \textbf{Time} & \textbf{Space} & \textbf{Biology} & \textbf{Physics} & \textbf{Chemistry} & \textbf{Overall} \\ \midrule

\multicolumn{9}{c}{\textit{Generation-only Models}} \\
\midrule
FLUX.1 [Dev] \cite{flux} & 12B & 0.48 & 0.58 & 0.62 & 0.42 & 0.51 & 0.35 & 0.50 \\
SD3.5-Large \cite{esser2024scaling} & 8B & 0.44 & 0.50 & 0.58 & 0.44 & 0.52 & 0.31 & 0.46 \\ \midrule

\multicolumn{9}{c}{\textit{Composite Unified Multimodal Models}} \\
\midrule
Blip3-o-4B \cite{blip3-o} & 4B & - & - & - & - & - & - & 0.50 \\
\midrule

\multicolumn{9}{c}{\textit{Smaller Native Unified Multimodal Models}} \\
\midrule
Harmon \cite{wu2025harmonizing} & 1.5B & 0.38 & 0.48 & 0.52 & 0.37 & 0.44 & 0.29 & 0.41 \\
Show-o2 \cite{xie2025show} & 1.5B & 0.33 & 0.41 & 0.53 & 0.35 & 0.49 & 0.31 & 0.39 \\
\rowcolor{icmlblue}
\model & 1.5B & \textbf{0.49} & \textbf{0.47} & \textbf{0.64} & \textbf{0.42} & \textbf{0.59} & \textbf{0.34} & \textbf{0.49} \\ \midrule

\multicolumn{9}{c}{\textit{Larger Native Unified Multimodal Models}} \\
\midrule
BAGEL \cite{bagel} & 14B & 0.44 & 0.55 & 0.68 & 0.44 & 0.60 & 0.39 & 0.52 \\
VILA-U-7B \cite{wu2024vila} & 7B & 0.26 & 0.33 & 0.37 & 0.35 & 0.39 & 0.23 & 0.31 \\
Janus-Pro-7B \cite{chen2025janus} & 7B & 0.30 & 0.37 & 0.49 & 0.36 & 0.42 & 0.26 & 0.35 \\
Emu3-Gen-8B \cite{wang2024emu3} & 8B & 0.34 & 0.45 & 0.48 & 0.41 & 0.45 & 0.27 & 0.39 \\
Show-o2 \cite{xie2025show} & 7B & 0.40 & 0.45 & 0.58 & 0.39 & 0.53 & 0.34 & 0.44 \\
\rowcolor{icmlblue}
\model & 7B & \textbf{0.52} & \textbf{0.53} & \textbf{0.68} & \textbf{0.47} & \textbf{0.58} & \textbf{0.38} & \textbf{0.53} \\ \bottomrule
\end{tabular}
\end{adjustbox}
\end{table*}

\begin{table}[!t]
\centering
\caption{Summary of licenses for existing assets used in this work.}
\label{tab:licenses}
\resizebox{0.85\textwidth}{!}{
\begin{tabular}{llc}
\toprule
\textbf{Asset Name} & \textbf{Asset Type} & \textbf{License} \\
\midrule
Qwen2.5 \cite{qwen2.5} & Pre-trained Model & Apache License 2.0 \\
InternViT \cite{internvl2.5} & Pre-trained Model & Apache License 2.0 \\
ImageNet-1K \cite{imagenet} & Dataset & Custom (Non-commercial research) \\
SA-1B  \cite{sam}& Dataset & Custom (Research use only) \\
\bottomrule
\end{tabular}
}
\end{table}
\section{Evaluation Details}
\label{appen:subsec:metric}

\subsection{Multi-modal Understanding Benchmarks}
To comprehensively evaluate the perception and reasoning capabilities of \model, we employ eight diverse benchmarks covering general understanding, expert knowledge, and fine-grained visual perception. We utilize {MME}~\cite{mme} to measure comprehensive perception and cognition, reporting {MME-S} as the aggregate sum of both scores. {MMBench (MMB)}~\cite{mmbench} is employed for its robust circular evaluation strategy, assessing fine-grained perception and logical reasoning. To ensure rational visual dependency, we evaluate on {MMStar}~\cite{chen2024mmstar}, a curated benchmark that filters out samples solvable by text alone. Additionally, {SEED-Bench (SEED)}~\cite{li2023seed} serves as a large-scale testbed for generative and discriminative comprehension across spatial and temporal dimensions. For high-level discipline-specific reasoning, we employ {MMMU}~\cite{yue2024mmmu}, a massive multi-discipline benchmark demanding expert-level reasoning across fields such as art, science, and engineering. This is complemented by {AI2D}~\cite{kembhavi2016ai2d}, a specialized dataset focused on understanding and answering questions about scientific diagrams and textbook illustrations. To assess text-centric visual understanding, we use {OCRBench}~\cite{liu2024ocrbench}, a dedicated benchmark for optical character recognition tasks. Furthermore, {RealWorldQA (RWQA)}~ is utilized to evaluate spatial reasoning and physical understanding in diverse real-world environments.

\section{Licenses for Existing Assets}
\label{app:licenses}

In our research, we utilize several open-source pre-trained models and datasets. We strictly adhere to their respective licenses and terms of use. The explicit licensing terms for the primary assets used in this paper are summarized in Table~\ref{tab:licenses}. 

\clearpage

\section{Qualitative comparison}
\subsection{Image Reconstruction}

\begin{figure}[h]
    \centering
    \includegraphics[width=0.9\linewidth]{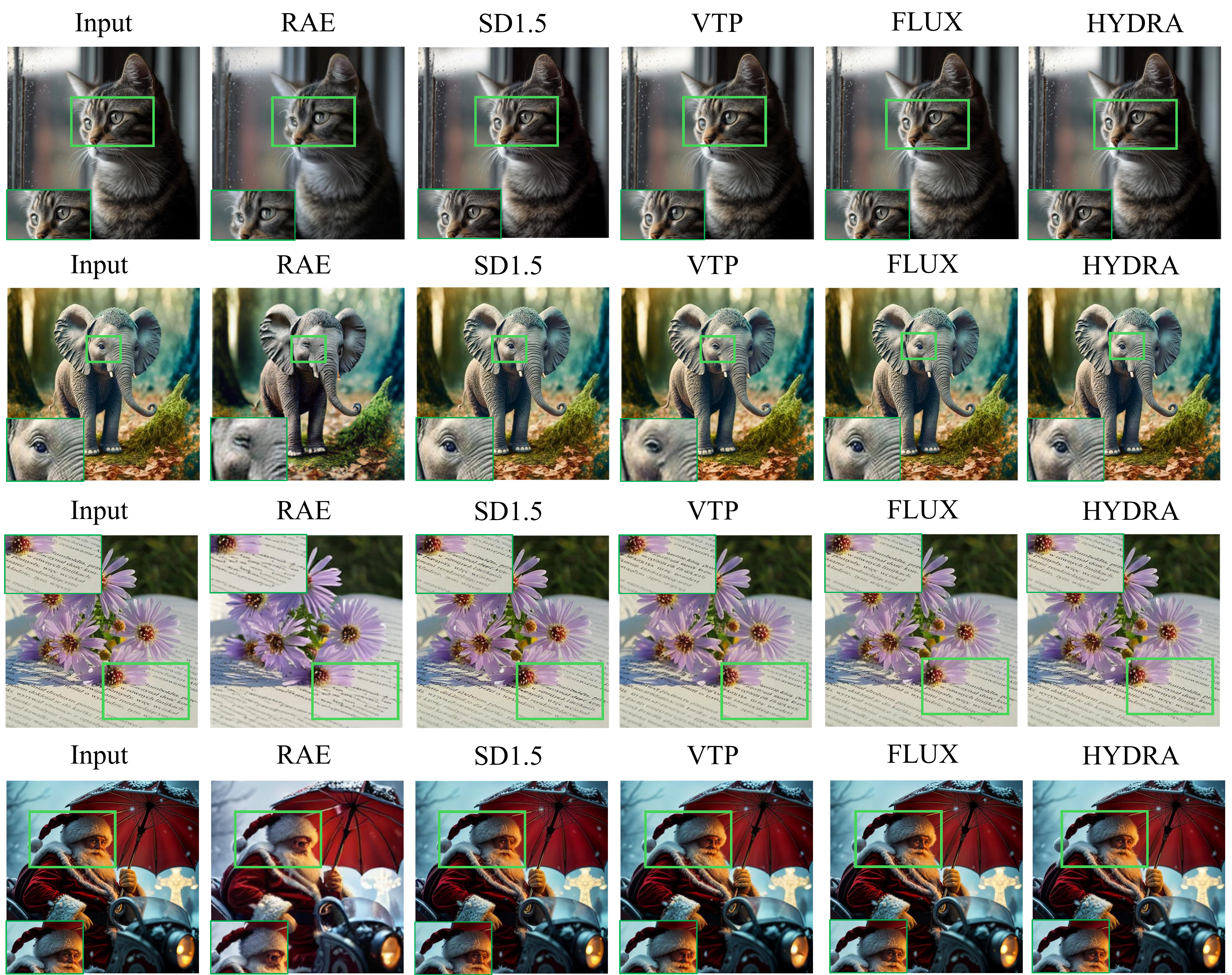}
    \caption{\textbf{Qualitative comparison of image reconstruction.} We compare visual results from our \model between RAE \cite{rae}, SD1.5 \cite{sdxl}, VTP \cite{yao2025vtp}, and FLUX \cite{flux}.}
    \label{fig:recon_compare}
\end{figure}
\clearpage

\subsection{Image Generation}
\label{vis}
\begin{figure}[h]
    \centering
    \includegraphics[width=0.87\linewidth]{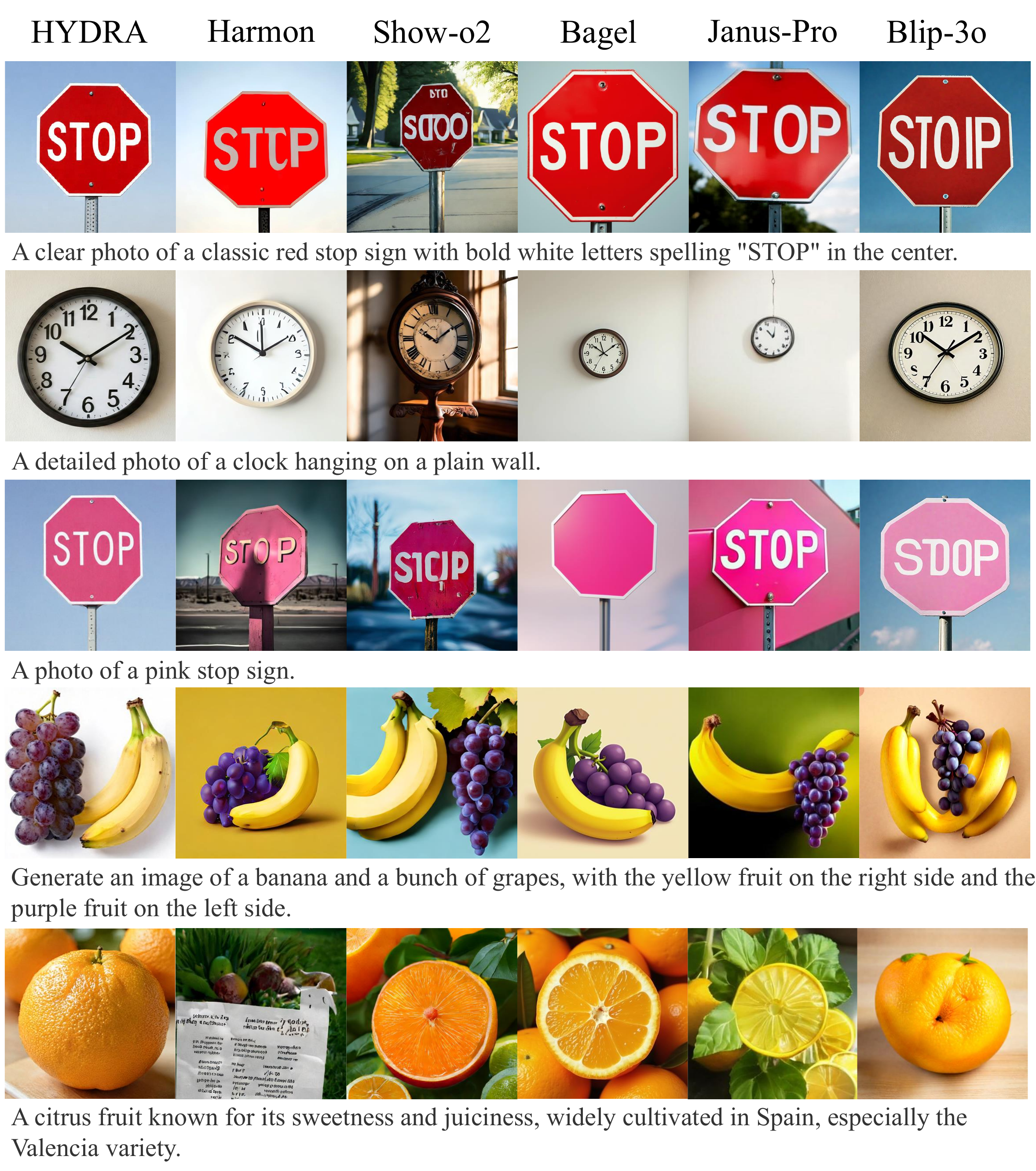}
    \caption{\textbf{Qualitative comparison of text-to-image generation.} We compare visual generation  between Harmon \cite{wu2025harmonizing}, Show-o2 \cite{xie2025show}, Bagel \cite{bagel}, Janus-pro \cite{wu2025janus}, and Blip-3o \cite{blip3-o}.}
    \label{fig:compare_gen1}
\end{figure}


\begin{figure}[h]
    \centering
    \includegraphics[width=0.87\linewidth]{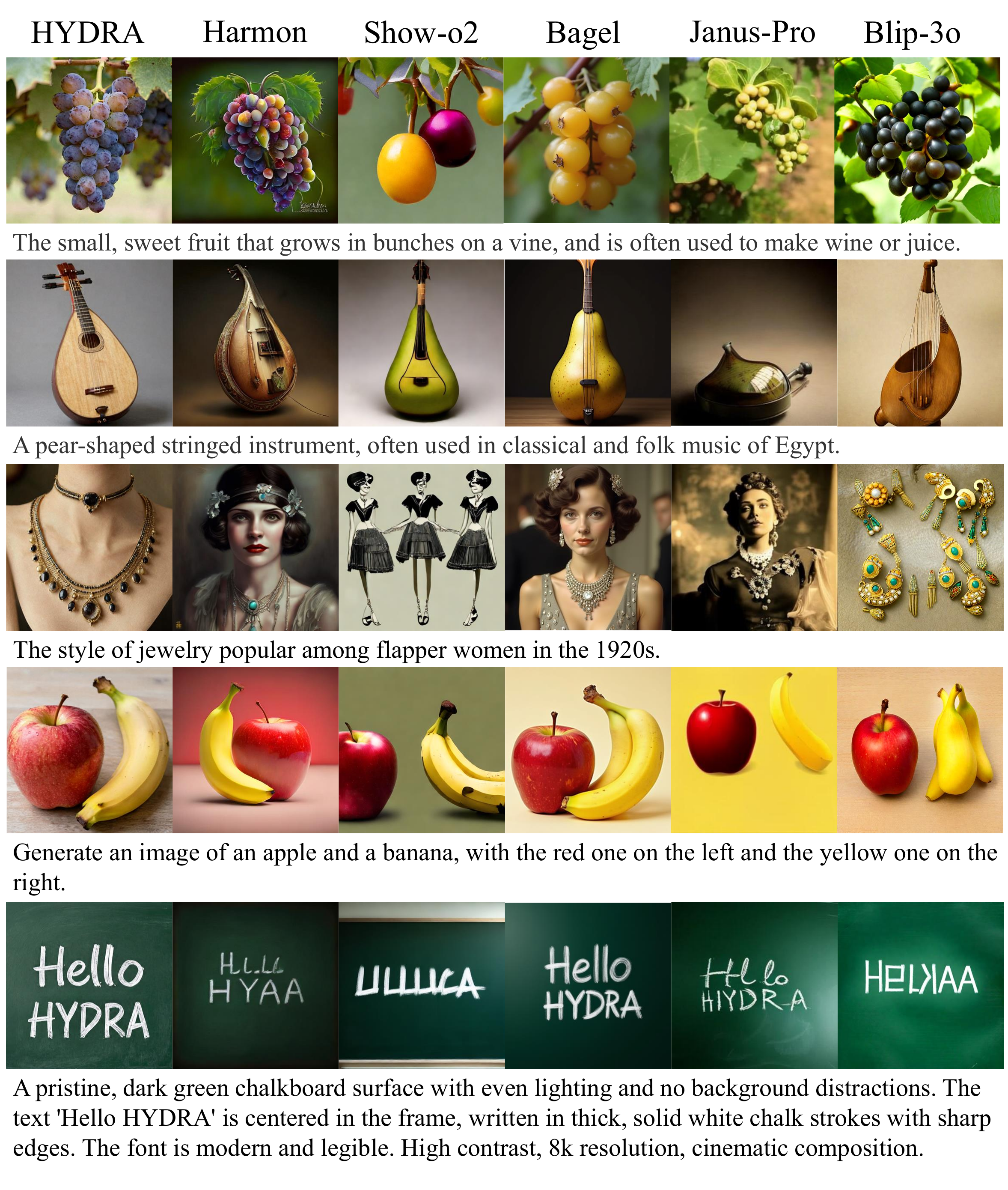}
    \caption{\textbf{Qualitative comparison of text-to-image generation.} We compare visual generation results from our \model between Harmon \cite{wu2025harmonizing}, Show-o2 \cite{xie2025show}, Bagel \cite{bagel}, Janus-pro \cite{wu2025janus}, and Blip-3o \cite{blip3-o}.}
    \label{fig:compare_gen2}
\end{figure}
\clearpage

\subsection{Multimodal Image Understanding}
\label{vis_mmu}
\begin{figure}[h]
    \centering
    \includegraphics[width=0.95\linewidth]{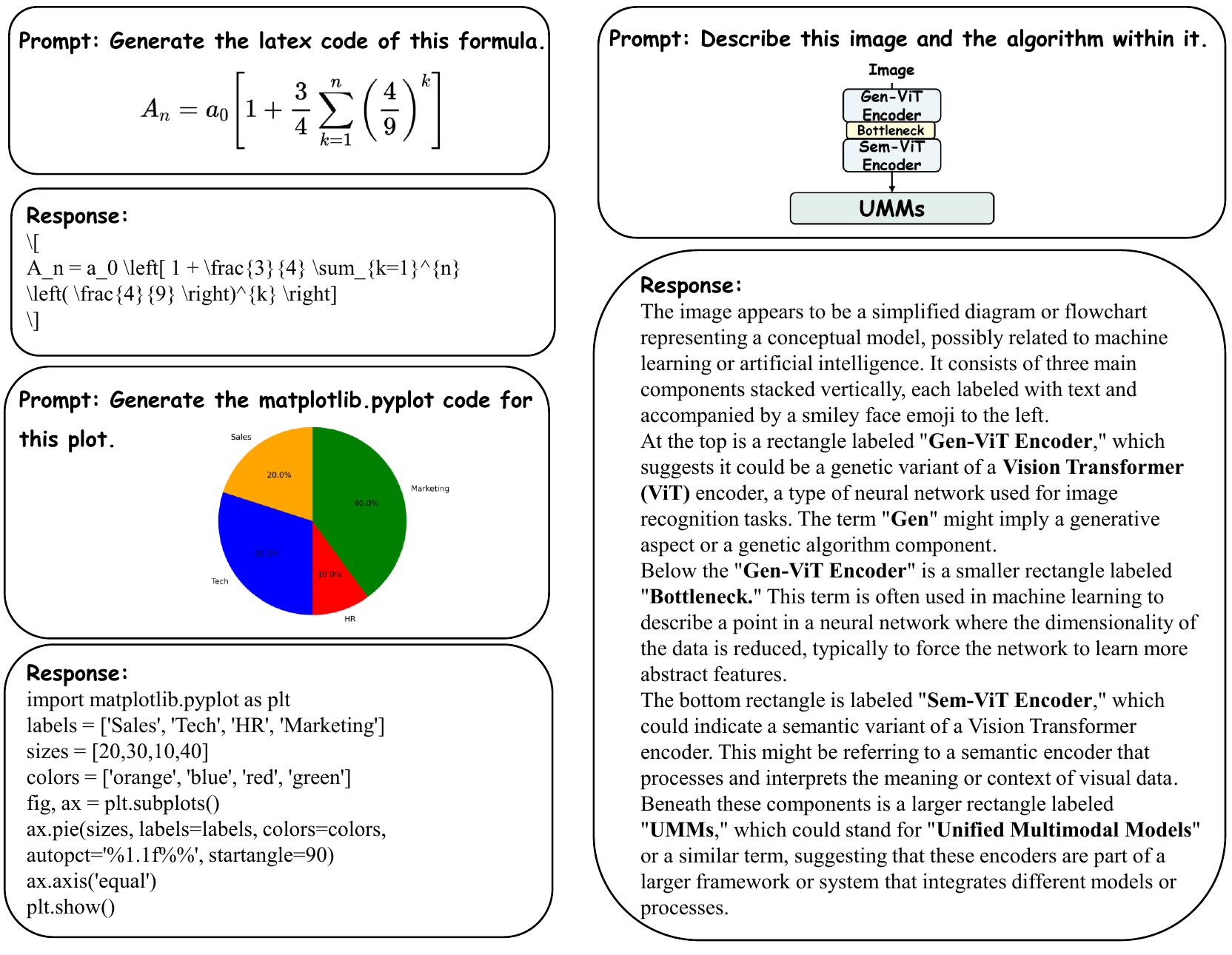}
    \caption{\textbf{Qualitative results on multimodal image Understanding.} First, \model exhibits superior proficiency in transforming complex visual data into structured text. From accurately generating LaTeX and Python code for formulas and charts to extracting fine-grained details from concept maps, the results highlight its powerful comprehension of diverse and abstract visual inputs.}
    \label{fig:compare_mmu}
\end{figure}

\newpage

\end{document}